\documentclass[letterpaper]{article} 
\usepackage[draft]{aaai2026}  
\usepackage{times}  
\usepackage{helvet}  
\usepackage{courier}  
\usepackage[hyphens]{url}  
\usepackage{graphicx} 
\usepackage{natbib}  
\usepackage{caption} 
\usepackage{algorithm}
\usepackage{algorithmic}

\usepackage{newfloat}
\usepackage{listings}
\DeclareCaptionStyle{ruled}{labelfont=normalfont,labelsep=colon,strut=off} 
\floatstyle{ruled}
\newfloat{listing}{tb}{lst}{}
\floatname{listing}{Listing}
\usepackage{amsmath,amsfonts,amssymb}
\usepackage{enumitem}
\usepackage{booktabs}
\usepackage{multirow}
\usepackage{subcaption}
\usepackage{tikz}
\newcommand{\mbf}[1]{\mbox{\boldmath $#1$}}
\DeclareMathOperator*{\argmin}{arg\,min}

\usepackage{color}

\title{FedSDA: Federated Stain Distribution Alignment for\\Non-IID Histopathological Image Classification}
\author{
    Cheng-Chang Tsai\textsuperscript{\rm 1,2},
    Kai-Wen Cheng\textsuperscript{\rm 1},
    Chun-Shien Lu\textsuperscript{\rm 1}
}
\affiliations{
    \textsuperscript{\rm 1}Institute of Information Science, Academia Sinica, Taipei, Taiwan, ROC\\
    \textsuperscript{\rm 2}Department of Computer Science and Information Engineering, National Taiwan University, Taipei, Taiwan, ROC\\

    cctsai831@gmail.com,
    kevin0606@iis.sinica.edu.tw,
    lcs@iis.sinica.edu.tw
}

\begin{document}

\maketitle

\begin{abstract}
    Federated learning (FL) has shown success in collaboratively training a model among decentralized data resources without directly sharing privacy-sensitive training data.
    Despite recent advances, non-IID (non-independent and identically distributed) data poses an inevitable challenge that hinders the use of FL.
    In this work, we address the issue of non-IID histopathological images with feature distribution shifts from an intuitive perspective that has only received limited attention.
    Specifically, we address this issue from the perspective of data distribution by solely adjusting the data distributions of all clients.
    Building on the success of diffusion models in fitting data distributions and leveraging stain separation to extract the pivotal features that are closely related to the non-IID properties of histopathological images, we propose a \textbf{Fed}erated \textbf{S}tain \textbf{D}istribution \textbf{A}lignment~(FedSDA) method.
    FedSDA aligns the stain distribution of each client with a target distribution in an FL framework to mitigate distribution shifts among clients.
    Furthermore, considering that training diffusion models on raw data in FL has been shown to be susceptible to privacy leakage risks, we circumvent this problem while still effectively achieving alignment.
    Extensive experimental results show that FedSDA is not only effective in improving baselines that focus on mitigating disparities across clients’ model updates but also outperforms baselines that address the non-IID data issues from the perspective of data distribution.
    We show that FedSDA provides valuable and practical insights for the computational pathology community.
\end{abstract}


\section{Introduction}
\begin{figure}[ht]
    \centering
    \begin{subfigure}[t]{0.3\textwidth}
        \centering\includegraphics[width=\linewidth]{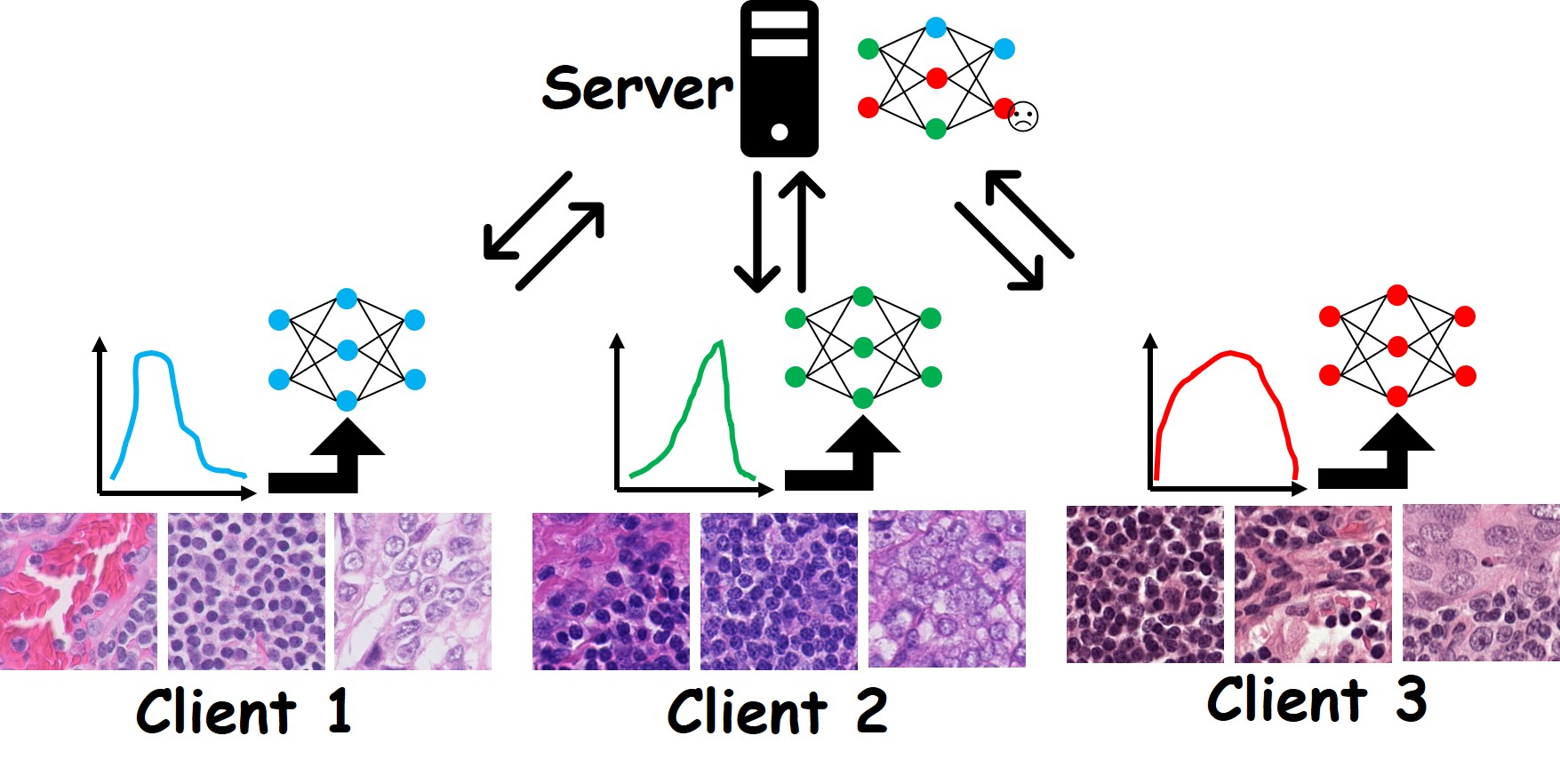}\caption{Inter-client feature distribution shifts}\label{fig:before_alignment}
    \end{subfigure}
    \hfill
    \begin{subfigure}[b]{0.3\textwidth}
        \centering\includegraphics[width=\linewidth]{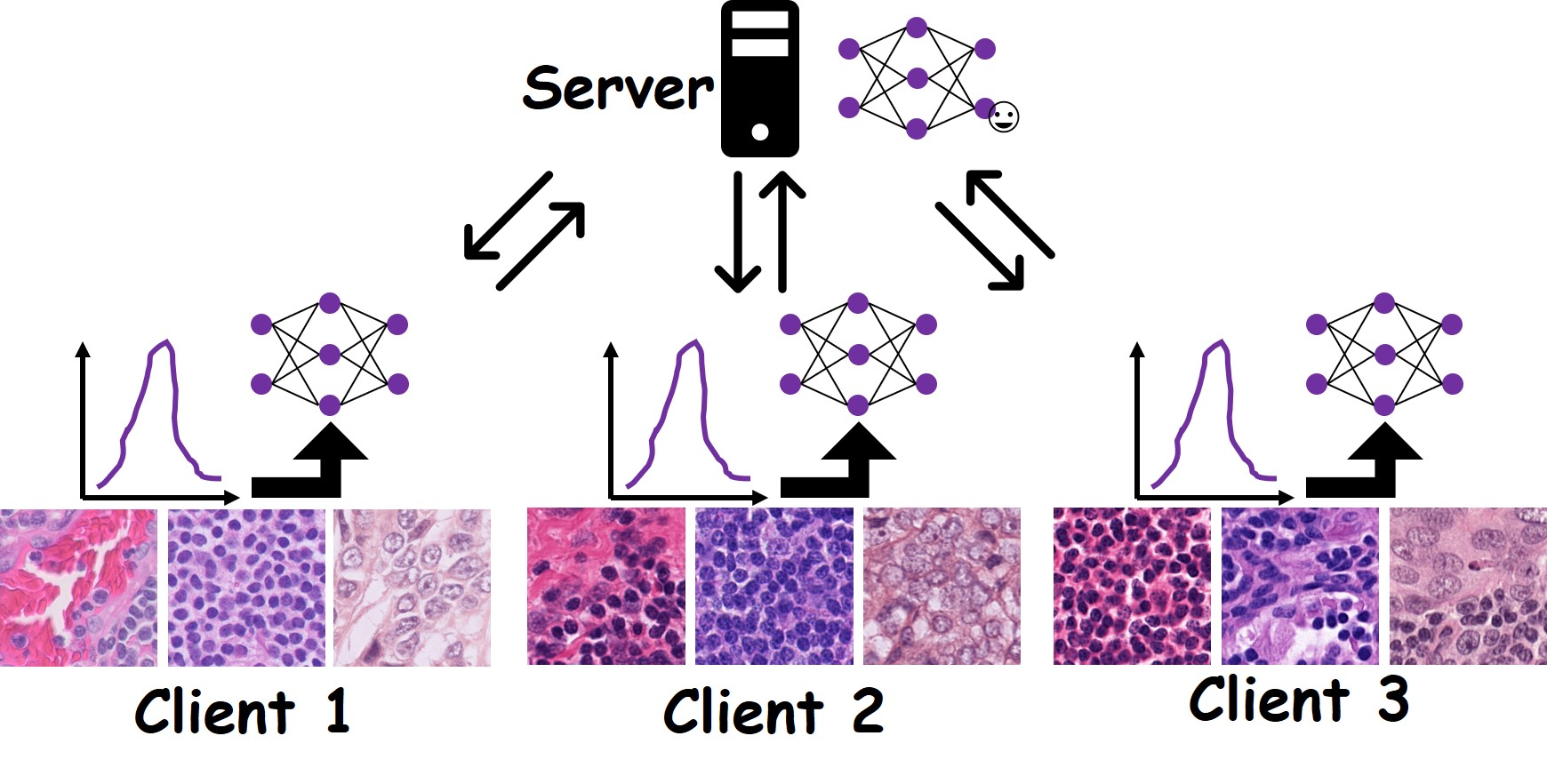}\caption{Mitigating distribution shifts via stain alignment}\label{fig:after_alignment}
    \end{subfigure}
    \caption{Illustration of two scenarios in FL. A model is collaboratively trained on histopathological images from multiple medical centers in a federated learning setting.}
    \label{fig:before_after}
\end{figure}
    \subsection{Background}
        In recent years, advances in machine learning~(ML) have significantly improved the performance of computer-aided diagnosis~(CAD) systems~\cite{chan2020computer,maleki2023breast,majumdar2023gamma,tsai2025test,raswa2025histofs}, which aid clinicians in analyzing medical images (e.g., histopathological images) and making more accurate and efficient decisions.
        Histopathological images play an indispensable role in the CAD of cancerous diseases~\cite{6857992,8854180,chen2017computer}.
        However, due to noise, missing values, rarity of certain diseases, cost issues, and privacy concerns, the available data in a single medical center for training an ML model is often limited, thereby affecting the performance of the trained model.
        To enhance the potential of data-driven ML models, collecting data from multiple medical centers is an intuitive and practical approach to increase the volume of data available for training models.
        However, the sensitive nature of medical images hinders the adoption of this approach.

        To address this challenge and facilitate training, federated learning~(FL)~\cite{mcmahan2017communication}, an emerging collaborative training framework that enables training on decentralized data from multiple medical centers without directly sharing sensitive training data, demonstrates great promise in addressing this hindrance.
        Despite the success of FL in histopathological image analysis~\cite{lu2022federated,adnan2022federated}, non-IID (non-independent and identically distributed) data among clients still significantly deteriorates the performance of FL systems~\cite{kairouz2021advances}.
        This issue is particularly prevalent in FL settings that involve training models on histopathological image data from multiple medical centers~\cite{8447230c17}, as depicted in Fig.~\ref{fig:before_alignment}.
        Therefore, this issue warrants urgent attention.
        In this work, we focus on non-IID data with feature distribution shifts (unless otherwise specified, non-IID data and distribution shifts refer to non-IID histopathological image data and feature distribution shifts, respectively)~\cite{li2021fedbn,zhou2023fedfa,qu2022generalized} rather than on non-IID data with label distribution shifts~\cite{deng2023scale,fediic2023wu}.
        
        Here is a commonly encountered example in a real medical center setting.
        Consider a model trained on data from one medical center to detect cancerous lesions in histopathological images.
        However, when deployed in another medical center, the model struggles due to differences in staining protocols and image acquisition scanners, which cause substantial chromatic variations across histopathological images~\cite{gupta2020gcti,tellez2019quantifying}.
        Even within the same medical center, it is common for a scanner of a different brand to be purchased, which can cause the model to not perform well in the histopathological images produced by the new scanner.
        This poses a significant challenge that hinders AI models from being portable and widely adopted in practical use.
        
    \subsection{Motivation}
        Existing works primarily address non-IID data issues in FL from the following two perspectives: (\emph{i}) Optimization~\cite{li2020federated,acar2021federated,gao2022feddc,qu2022generalized,pmlr-v235-fan24c,FedGAMMA,karimireddy2020scaffold} and (\emph{ii}) Normalization~\cite{li2021fedbn,zhangunderstanding,kang2024fednn,wagner2022federated,shen2022federated}.
        Despite these efforts, little attention has been paid to addressing non-IID data issues in FL solely from the perspective of data distribution (i.e., adjusting the data distributions of all clients to mitigate non-IID data issues, as depicted in Fig.~\ref{fig:after_alignment}).
        Among the limited works in this direction, CCST~\cite{chen2023federated} is the one specifically proposed to address non-IID data issues in FL by aligning each client's data distribution with those of all the other.

        However, we have observed that CCST~\cite{chen2023federated} cannot preserve the structural information of histopathological images well, as shown in Figs.~\ref{fig:c17_after_processing} and~\ref{fig:a22_after_processing}  (in the Appendix).
        This motivates us to design a better method that minimizes the loss of structural information.
        Since the core issue of non-IID data lies in stain variations, a promising approach is to separate histopathological images into the part containing stain-related information and the part containing structural information and to modify only the stain-related part while keeping the structural part unchanged.
        To achieve such separation, we adopt stain separation~\cite{vahadane}.
        Then, we adjust the distributions of the stain matrices to align the data distributions with each other, thereby addressing the non-IID data issues.
        However, in FL, stain separation~\cite{vahadane} cannot reconstruct histopathological images for alignment without access to the stains from the other clients.
        To address this issue, we adopt a diffusion model~\cite{ho2020denoising} trained using FedAvg~\cite{mcmahan2017communication} to fit the data distribution collectively formed by all clients (i.e., target distribution).
        Although this approach can achieve the goal, recent studies~\cite{zhu2025fedmia,gan2024datastealing} have shown that training diffusion models on raw data via FedAvg carries a high risk of privacy leakage.
        For example, communicated gradients have been shown to leak private data under certain training configurations~\cite{zhu2019deep}.
        Moreover, FedMIA~\cite{zhu2025fedmia} can effectively determine whether a specific data sample belongs to the private dataset.
        However, among these threats, DataSteal~\cite{gan2024datastealing} is a method specifically designed for diffusion models trained via federated learning and demonstrates superior attack performance.
        To mitigate these privacy risks, we train the diffusion model on stain matrices rather than raw data.

        With all these concerns adequately addressed, we propose \textbf{Fed}erated \textbf{S}tain \textbf{D}istribution \textbf{A}lignment (FedSDA), a method that aligns the stain distributions of all clients in an FL system with each other, as depicted in Fig.~\ref{fig:after_alignment}.
        It is noteworthy that FedSDA shares a similar goal with normalization-based methods~\cite{jiang2022harmofl,li2021fedbn}.
        However, FedSDA has the advantage of being aware of the structure of histopathological images and the stain distribution of data, but normalization-based methods do not pay special attention to these aspects.
        Furthermore, FedSDA does not require additional processing during testing, whereas normalization-based methods do.

    \subsection{Contributions}
        Our contributions are summarized as follows:
        \begin{itemize}
            \item We propose a federated stain distribution alignment method that mitigates feature distribution shifts using a diffusion model trained via federated learning, enabling each client to access the stains of all the other clients.
            \item The proposed method circumvents the need to train the diffusion model on raw data while still effectively achieving alignment, as direct training on raw data in FL has been shown to be susceptible to privacy leakage risks.
            \item Extensive experimental comparisons demonstrate that the proposed method not only improves the classification performance of the baselines~\cite{li2020federated,qu2022generalized,pmlr-v235-fan24c} that focus on mitigating disparities across clients’ model updates but also outperforms those~\cite{chen2023federated,jiang2022harmofl,zhou2023fedfa} that address non-IID data issues from the perspective of data distribution.
        \end{itemize}

\section{Related Work}
    \subsection{Federated Learning}
        FedAvg~\cite{mcmahan2017communication} is a seminal work in FL and can be regarded as the foundation of all subsequent works.
        \subsubsection{Optimization Perspective}
        The methods reviewed here share a similar design rationale to address non-IID data issues by mitigating disparities among updates.
        FedProx~\cite{li2020federated} incorporates a proximal term into the loss function during local training to reduce the discrepancy between local and global models.
        SCAFFORD~\cite{karimireddy2020scaffold} corrects local updates by estimating the difference between global and local update directions.
        FedDyn~\cite{acar2021federated} dynamically updates each client's regularizer in each round to align the local solution with the solution of the global loss function in the limit.
        FedDC~\cite{gao2022feddc} introduces an auxiliary variable in local training to track the gap between local and global models.
        FedSAM~\cite{qu2022generalized} leverages the Sharpness Aware Minimization (SAM)~\cite{foret2020sharpness} local optimizer to enhance local learning generalization.
        To further improve FedSAM~\cite{qu2022generalized}, FedLESAM~\cite{pmlr-v235-fan24c} locally estimates the global perturbation direction on the client side by computing the difference between global models received in the previous active round and the current round.
        FedGAMMA~\cite{FedGAMMA} incorporates FedSAM~\cite{qu2022generalized} with the correction technique proposed by SCAFFORD~\cite{karimireddy2020scaffold} to mitigate disparities across all client updates.
        \subsubsection{Normalization Perspective}
            In contrast, the methods reviewed here employ various normalization techniques to mitigate distribution shifts.
            However, these normalization techniques often induce the data to conform to a single feature rather than retaining the original distribution of the data (e.g., in Figs.~\ref{fig:c17_after_processing} and~\ref{fig:a22_after_processing} in the Appendix, the images in the bottom row are normalized to a similar stain, whereas those in the second row from the top exhibit more diverse stains).
            This highlights the fundamental difference between the proposed method and normalization-based methods.
            FedBN~\cite{li2021fedbn}, built on FedAvg~\cite{mcmahan2017communication}, uses batch normalization~(BN) layers~\cite{ioffe2015batch} locally and excludes the parameters of the BN layers from aggregation on the central server.
            \citeauthor{zhangunderstanding} study the role of layer normalization~(LN)~\cite{ba2016layer} on non-IID data with label distribution shifts in FL, and their results verify that applying normalization to the feature from a classifier is the essence of LN, which improves classification performance under extreme label distribution shifts.
            FedNN~\cite{kang2024fednn} uses weight normalization~(WN)~\cite{salimans2016weight} and adaptive group normalization~(AGN) to mitigate variations in model weights and features, respectively, in concept drift data (i.e., feature distribution shifts).
            Finally, HarmoFL~\cite{jiang2022harmofl} employs SAM~\cite{foret2020sharpness} and amplitude normalization to mitigate local and global drift, respectively.
    \subsection{Distribution Shifts}
        Non-IID data poses a significant issue in FL, also known as distribution shifts~\cite{gao2023back,tsai2024gda,zhang2021adaptive,koh2021wilds,raswa2025histofs}, where gaps exist among data distributions.
        Many previous works~\cite{gao2023back,tsai2024gda,nie2022diffusion,tsai2025test,hoffman2018cycada} have been proposed to address this issue from the perspective of domain adaptation using generative models.
        Specifically, they adapt data from one dataset (i.e., target domain) to match the distribution of another dataset (i.e., source domain).
        CyCADA~\cite{hoffman2018cycada}, based on CycleGAN~\cite{zhu2017unpaired}, adapts data by translating between domains, guided by a specific discriminatively trained task.
        DiffPure~\cite{nie2022diffusion} uses diffusion models~\cite{ho2020denoising} to purify adversarial images~\cite{goodfellow2014explaining} (i.e., images from target domain), generating purified images that closely follow the distribution of clean images (i.e., images from source domain).
        DDA~\cite{gao2023back} also uses diffusion models~\cite{ho2020denoising} to adapt images from a target domain (or possibly multiple target domains) to a source domain without significantly losing class information, considering the distance between the reference and noisy images after applying a low-pass filter.
        GDA~\cite{tsai2024gda} refines DDA~\cite{gao2023back} by adopting the structural guidance, which is defined by a marginal entropy loss derived from the classifier, along with style and content preservation losses.
        TT-SaD~\cite{tsai2025test} adapts the stains of histopathological images using diffusion models~\cite{ho2020denoising}.
        In addition to these domain adaptation methods that are not designed for FL,~\citeauthor{chen2023federated} propose CCST~\cite{chen2023federated}, which solely adjusts the data distributions of all clients by augmenting the data in each client through cross-client style transfer based on adaptive instance normalization (AdaIN)~\cite{huang2017arbitrary} for federated domain generalization.
\section{Preliminaries}
    \subsection{Notations}
        In an FL system of $K$ clients, the $i$-th client, denoted by $C_{i}$, only has access to histopathological images $\mathcal{X}_{i} = \{ \mathbf{x}_{i,j} \}^{n_{i}}_{j=1}$ and the corresponding labels $\mathcal{Y}_{i} = \{ \mathbf{y}_{i,j} \}^{n_{i}}_{j=1}$, and they constitute a dataset $\mathcal{D}_{i} = \{ ( \mathbf{x}_{i,j} , \mathbf{y}_{i,j} ) \}^{n_{i}}_{j=1}$, where $n_{i}$ denotes the number of data in $C_{i}$, for $i \in \{ 1 , 2 , \cdots , K \}$.
        In addition, the data distribution of $C_{i}$ is denoted by $\mathcal{P}_{i}$, and all data pairs $( \mathbf{x}_{i} , \mathbf{y}_{i} ) \in \mathcal{D}_{i}$ are drawn from $\mathcal{P}_{i}$, i.e., $( \mathbf{x}_{i} , \mathbf{y}_{i} ) \sim \mathcal{P}_{i}$.
        Finally, let $\mathbb{E}$, $\log ( \cdot )$, $\| \cdot \|_{F}$, $\| \cdot \|_{1}$, $\| \cdot \|_{2}$, $\mathbf{x} ( : , i )$, and $\exp{( \cdot )}$ denote the expectation, the element-wise logarithmic function, the Frobenius norm, the $\ell_1$ norm, the $\ell_2$ norm, the $i$-th column of $\mathbf{x}$, and the element-wise exponential function, respectively.
        We will frequently omit subscripts for simplicity. 
    \subsection{Problem Formulation}
        The goal of FL is to train a model $f_{\theta}$, parameterized by $\theta$, on decentralized data without directly sharing privacy-sensitive training data.
        Specifically, the objective can be formulated as follows~\cite{mcmahan2017communication,jiang2021unifed}: \begin{equation}\label{eq:FL_objective}
            \argmin_{\theta} \frac{1}{K} \sum^{K}_{i=1} \mathbb{E}_{( \mathbf{x} , \mathbf{y} ) \in \mathcal{D}_{i}} \left[ \ell \left( f_{\theta} ( \mathbf{x} ) , \mathbf{y} \right) \right],
        \end{equation}
        where $\ell$ denotes a task-dependent loss function, e.g., cross-entropy loss for a classification task.

        In this work, we consider solving Eq.~(\ref{eq:FL_objective}) under inter-client feature distribution shifts~\cite{tan2024heterogeneity}, which are defined as follows:
        \begin{equation}\label{eq:inter_client_feature_distribution_shift}
            \mathcal{P}_{i} ( \mathbf{x} | \mathbf{y} ) \neq \mathcal{P}_{j} ( \mathbf{x} | \mathbf{y} ) \quad \mbox{for all $i \neq j$}.
        \end{equation}
        This scenario is prevalent in histopathological image analysis~\cite{8447230c17,wilm2023multiscanner}, as depicted in Fig.~\ref{fig:before_alignment}.
\begin{figure}[ht]
    \centering\includegraphics[width=0.65\linewidth]{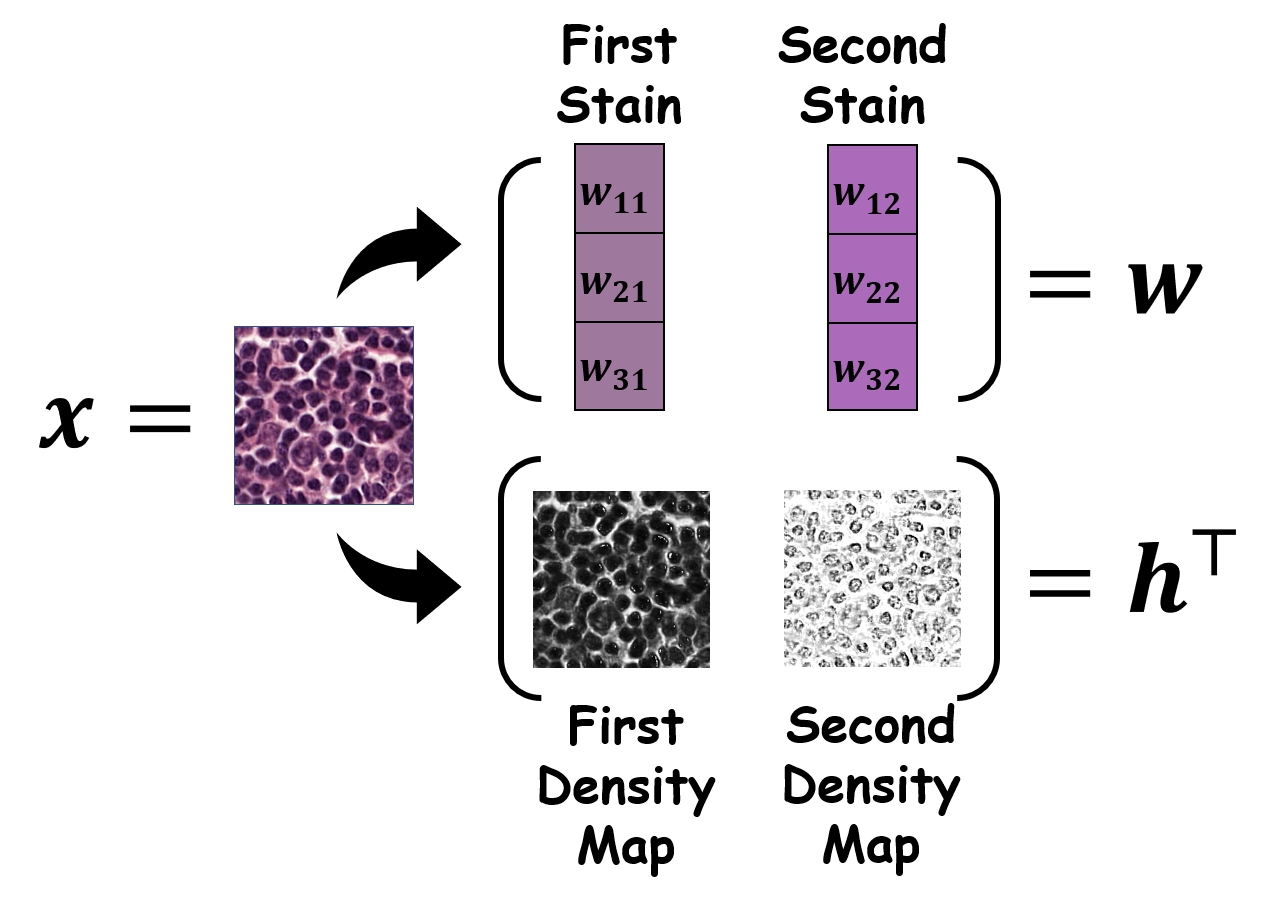}\caption{Illustration of stain separation. Given an H{\&}E-stained histopathological image $\mathbf{x}$, we can decompose it into a stain matrix $\mathbf{w}$ of size $3 \times 2$ and a stain density map $\mathbf{h}$ of size $2 \times N$, where $r$ represents the number of stains (in this case, $r = 2$). Each $\mathbf{w}_{ij}$ represents an element of $\mathbf{w}$.}\label{fig:stain_separation}
\end{figure}
    \subsection{Stain Separation}
        Stain separation~\cite{vahadane}, as depicted in Fig.~\ref{fig:stain_separation}, aims to decompose a histopathological image $\mathbf{x} \in \mathbb{R}^{m \times N}$, where $m=3$ denotes the RGB channels and $N$ is the number of pixels in $\mathbf{x}$, into a stain matrix, denoted by $\mathbf{w} \in \mathbb{R}^{m \times r}$, and the corresponding stain density map, denoted by $\mathbf{h} \in \mathbb{R}^{r \times N}$, so that $\mathbf{x}$ can be reconstructed by $\mathbf{w}$ and $\mathbf{h}$.
        Note that the columns of $\mathbf{w}$ and the rows of $\mathbf{h}$ represent the chromatic basis of each stain and the concentration of each stain, respectively.
        For more details on stain separation, please refer to~\cite{vahadane}.

        Stain separation is formulated as the following optimization problem:
        \begin{align}\label{eq:stain_separation_optimization}
            \argmin_{\mathbf{w} , \mathbf{h}} & \mbox{ } \frac{1}{2} \left\| - \log \frac{\mathbf{x}}{I_0} - \mathbf{w} \mathbf{h} \right\|^{2}_{F} + \lambda \left\| \mathbf{h} \right\|_{1}\\
            \mbox{s.t.} & \quad \mathbf{w} , \mathbf{h} \geq 0 \quad \mbox{and} \quad \left\| \mathbf{w} ( : , i ) \right\|^{2}_{2} = 1 \mbox{ } \forall i , \nonumber
        \end{align}
        where $\lambda$ is a regularization parameter that balances the regularization and loss terms, and $I_{0}$ is the illuminating light intensity of an image (typically 255 for 8-bit images).
        This optimization problem is solved iteratively by optimizing one set of parameters (e.g., $\mathbf{w}$) while keeping the other (i.e., $\mathbf{h}$) fixed.
        After obtaining the stain matrix $\mathbf{w}$ and the stain density map $\mathbf{h}$ of a histopathological image $\mathbf{x}$, the reconstructed image $\hat{\mathbf{x}}$ is computed as:
        \begin{equation}\label{eq:reconstruction}
            \hat{\mathbf{x}} = I_{0} \exp{( - \mathbf{w} \mathbf{h})}.
        \end{equation}
        For clarity, we denote Eq.~(\ref{eq:stain_separation_optimization}) and Eq.~(\ref{eq:reconstruction}) as $\mathcal{S}$ and $\mathcal{R}$, respectively, which leads to the following two expressions:
        \begin{equation}\label{eq:SR}
            \{ \mathbf{w} , \mathbf{h} \} = \mathcal{S} ( \mathbf{x} ) \quad \mbox{and} \quad \hat{\mathbf{x}} = \mathcal{R}( \mathbf{w} , \mathbf{h} ).
        \end{equation}
    \subsection{Diffusion Models}
        Diffusion models~\cite{ho2020denoising} are known to fit a given data distribution, denoted by $q ( \mathbf{x}_{0} )$, so that the generated data belong to $q ( \mathbf{x}_{0} )$, and consist of two processes (i.e., the forward and reverse processes).
        Given $\mathbf{x}_{0} \sim q ( \mathbf{x}_{0} )$, the forward process gradually adds Gaussian noise to $\mathbf{x}_{0}$ to produce latent samples $\mathbf{x}_{1} , \mathbf{x}_{2} , \ldots , \mathbf{x}_{T}$, where $T$ is a predefined constant, via
        \begin{equation}\label{eq:forward}
            q ( \mathbf{x}_{t} | \mathbf{x}_{t-1} ) := \mathcal{N} \left( \mathbf{x}_{t} ; \sqrt{1 - \beta_{t}} \mathbf{x}_{t-1} , \beta_{t} \mathbf{I} \right),
        \end{equation}
        where $\{ \beta_{t} \}^{T}_{t=1}$ is a variance schedule, $\mathcal{N}$ denotes a Gaussian distribution, and $\mathbf{I}$ denotes the identity matrix.
        Conversely, the reverse process is formulated as:
        \begin{equation}\label{eq:reverse}
            p_{\theta} ( \mathbf{x}_{t-1} | \mathbf{x}_{t} ) := \mathcal{N} \left( \mathbf{x}_{t-1} ; \frac{\mathbf{x}_{t}}{\sqrt{\alpha_{t}}} - \frac{\beta_{t}\epsilon_{\theta} ( \mathbf{x}_{t} , t )}{\sqrt{\alpha_{t} ( 1 - \bar{\alpha}_{t} )}} , \beta_{t} \mathbf{I} \right),
        \end{equation}
        where $\alpha_{t} := 1 - \beta_{t}$, $\bar{\alpha}_{t} := \prod^{t}_{s=1} \alpha_{s}$, and $\epsilon_{\theta}$ is a neural network parameterized by $\theta$.
        Finally, to train a conditional diffusion model, we optimize the following problem:
        \begin{equation}
            \min_{\theta} \mathbb{E}_{t , \mathbf{x}_{0} , c , \mathbf{\epsilon}} \left\| \mathbf{\epsilon} - \epsilon_{\theta} ( \mathbf{x}_{t} , t , c ) \right\|^{2}_{2},
        \end{equation}
        where $t$ is the timestep, $c$ is the condition, and $\mathbf{\epsilon} \sim \mathcal{N} ( 0 , \mathbf{I} )$.
\begin{figure*}[ht]
    \centering\includegraphics[width=0.88\linewidth]{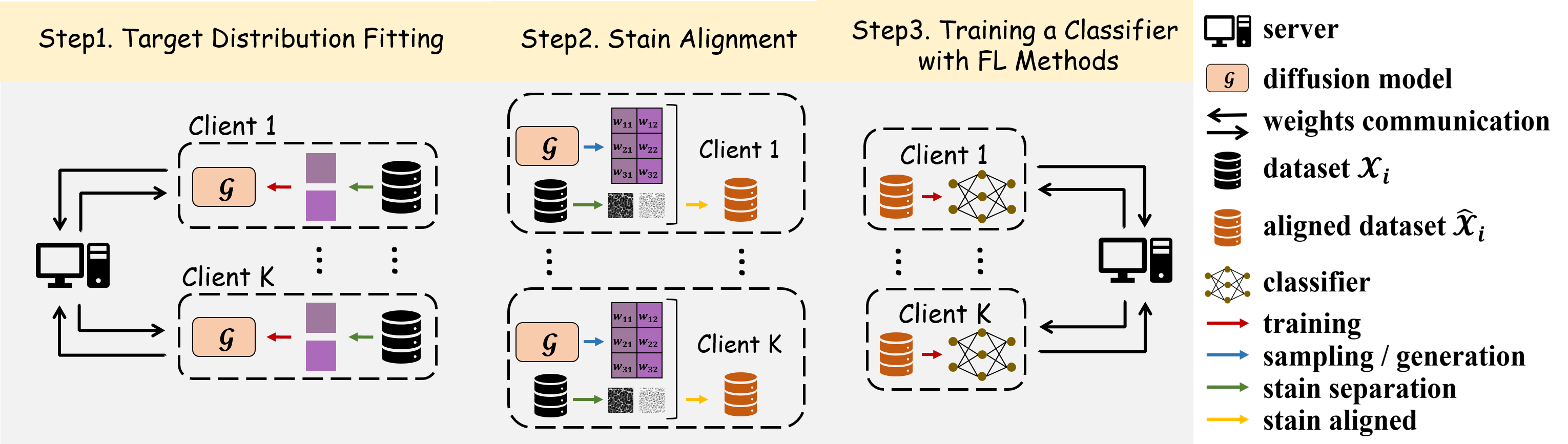}\caption{Workflow of our FedSDA method}\label{fig:workflow}
\end{figure*}
\begin{figure}[ht]
    \centering\includegraphics[width=0.55\linewidth]{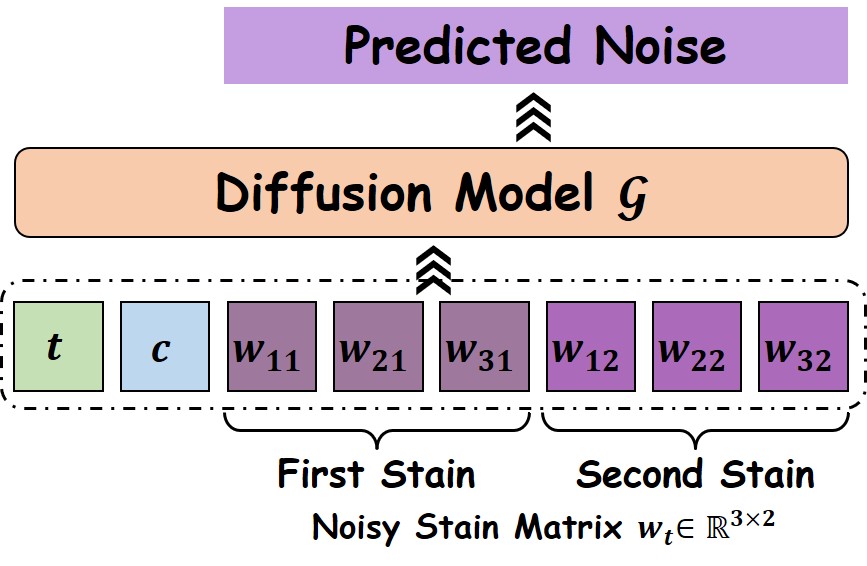}\caption{Illustration of how $\mathbf{w}_{t}$ is fed into $\mathcal{G}$.}\label{fig:uvit_input}
\end{figure}
\section{Method}
    \subsection{Overview of FedSDA}
        To mitigate the issues defined in Eq.~(\ref{eq:inter_client_feature_distribution_shift}), FedSDA, which is proposed to align the stain distributions of all clients with a target distribution, consists of two steps: (\emph{i}) Target distribution fitting and (\emph{ii}) Stain alignment.
        The first step is to use a diffusion model to fit the stain distribution formed collectively by all clients (i.e., target distribution).
        This diffusion model provides each client with access to the stain matrices of the other clients, which is a necessary component for the second step.
        With the diffusion model, the second step can be applied to histopathological images of each client, thereby achieving the goal of addressing the non-IID data issues.
        We provide the workflow of FedSDA in Fig.~\ref{fig:workflow}.

    \subsection{Target Distribution Fitting}
        FedSDA requires a generative model to generate the stain matrices of the other clients for the second step of FedSDA (i.e., stain alignment).
        Due to the recent success of diffusion models in generating synthetic data and modeling complex distributions~\cite{kotelnikov2023tabddpm,dhariwal2021diffusion,ho2022video}, we adopt a diffusion model as the generative model for FedSDA.

        Here, we introduce the diffusion model adopted in FedSDA.
        Since the model is trained on low-dimensional data, we adopt a single-layer transformer model as the backbone of the diffusion model, denoted by $\mathcal{G}$.
        This lightweight network architecture ensures that the additional overhead is minimal.
        This model takes the timestep $t$, the condition $c$, and the noisy input $\mathbf{w}_{t}$ as inputs, and estimates the noise added to $\mathbf{w}_{t}$.
        Since the noisy data $\mathbf{w}_{t}$ fed into $\mathcal{G}$ is a stain matrix with added noise rather than a sequence of tokens, it is necessary to adjust the way $\mathbf{w}_{t}$ is processed by $\mathcal{G}$.
        Specifically, we need to determine how $\mathbf{w}_{t}$ is represented as tokens, as $\mathcal{G}$ treats all inputs, including $t$, $c$, and $\mathbf{w}_{t}$, as tokens.
        Intuitively, we treat each element of $\mathbf{w}_{t}$ as a token and concatenate $(t,c)$ with the tokens of $\mathbf{w}_{t}$ to form the input of $\mathcal{G}$, as depicted in Fig.~\ref{fig:uvit_input}.
        
        To train a diffusion model on stain matrices from all clients without directly sharing them, we follow the previous works~\cite{huang2024gradient,tun2023federated,vora2024feddm,de2024training} and adopt FedAvg~\cite{mcmahan2017communication}.
        Specifically, in each communication round, each client downloads the diffusion model weights from the central server, then trains the model locally, and finally uploads the local weights to the central server for aggregation.
        We provide the pseudocode in Algorithm~\ref{alg:train_cdm} (in the Appendix) for this training process.
        Note that the diffusion model we train is conditional, meaning that each stain matrix requires a condition $c$ to form an input for $\mathcal{G}$.
        For client $C_{i}$, we use its index $i$ as the condition $c$ for the corresponding stain matrices.
        For ease of reference, we denote the stain matrix generated under a given condition $c$ as $\mathcal{G} ( c )$.
        The condition used here determines the client to which the stain matrices belong and corresponds to the client index.
        Therefore, to generate the stain matrices belonging to client $C_{i}$, we apply $\mathcal{G} ( i )$.             
        
    \subsection{Stain Alignment}
        In the first step of FedSDA, the central server ends up with a conditional diffusion model $\mathcal{G}$, trained on the stain matrices from all clients in an FL system.
        This diffusion model $\mathcal{G}$ is used in the second step of FedSDA to sample stain matrices belonging to other clients for stain alignment.
        With access to other clients' stain matrices, we can align the data distribution of client $C_{i}$ with some distribution $\hat{\mathcal{P}}$ (ideally, it is the data distribution considering all clients' data as a whole), i.e., $\mathcal{P}_{i} ( \mathbf{x} | \mathbf{y} ) \approx \hat{\mathcal{P}} ( \mathbf{x} | \mathbf{y} ).$
        As a result of aligning through all clients, we can obtain the following results for clients $C_{i}$ and $C_{j}$:
        \begin{equation}\label{eq:all_align}
            \mathcal{P}_{i} ( \mathbf{x} | \mathbf{y} ) \approx \mathcal{P}_{j} ( \mathbf{x} | \mathbf{y} ) \quad \mbox{for all $i \neq j$},
        \end{equation}
        which mitigates the non-IID characteristic of data across different clients.
        Note that the reason we can achieve Eq.~(\ref{eq:all_align}) is that we have identified stain as the essential factor that causes feature distribution shifts.
        Therefore, reducing stain distribution shifts can achieve Eq.~(\ref{eq:all_align}).

        Specifically, the diffusion model $\mathcal{G}$ is downloaded by all clients from the central server.
        Since all clients follow the same procedure (denoted by $\mathcal{A}$ hereafter) to achieve Eq.~(\ref{eq:all_align}), we present the procedure $\mathcal{A}$ for a given client $C_{i}$ as an example for simplicity, and the same procedure applies to all other clients.
        We provide the pseudocode in Algorithm~\ref{alg:stain_alignment} (in the Appendix).

        Given a dataset $\mathcal{D}_{i}$ of client $C_{i}$, due to the label-free nature of FedSDA, $\mathcal{A}$ only requires the input dataset $\mathcal{X}_{i}$ to obtain an aligned dataset $\hat{\mathcal{X}_{i}}$, i.e., $\hat{\mathcal{X}_{i}} = \mathcal{A} ( \mathcal{X}_{i} )$.
        To balance the stains from each client in $\hat{\mathcal{X}_{i}}$, we randomly partition $\mathcal{X}_{i}$ into $K$ (i.e., the number of clients) subsets as $\mathcal{X}_{i} = \left\{ \mathcal{X}_{i,1} , \mathcal{X}_{i,2} , \cdots , \mathcal{X}_{i,K} \right\}$, and each subset contains the same number of samples (denoted by $n = n_{i} / K$).
        For each subset indexed by $j$ ($1\leq j\leq K$), we use $\mathcal{S}$ (in Eq.~(\ref{eq:SR})) to obtain the corresponding stain density maps (denoted by $\left\{ \mathbf{h}_{i,j,k} \right\}^{n}_{k = 1}$ hereafter) from the input data (denoted by $\left\{ \mathbf{x}_{i,j,k} \right\}^{n}_{k = 1}$ hereafter) by applying $\mathcal{S} \left( \mathbf{x}_{i,j,k} \right)$ for each $k$, where $\mathcal{X}_{i,j} = \left\{ \mathbf{x}_{i,j,k} \right\}^{n}_{k = 1}$. 
        Then, $\mathcal{R}$ (in Eq.~(\ref{eq:SR})) is applied to reconstruct a dataset with the stain of client $C_{j}$.
        Specifically, for each $k \in \left\{ 1 , 2 , \cdots , n \right\}$, the reconstructed $\hat{\mathbf{x}}_{i,j,k}$ is obtained by $\mathcal{R} \left( \mathcal{G} ( j ) , \mathbf{h}_{i,j,k} \right)$, i.e., $\hat{\mathbf{x}}_{i,j,k} = \mathcal{R} \left( \mathcal{G} ( j ) , \mathbf{h}_{i,j,k} \right)$.
        Thus, we obtain the reconstructed dataset $\hat{\mathcal{X}}_{i,j} = \left\{ \hat{\mathbf{x}}_{i,j,k} \right\}^{n}_{k = 1}$, and further obtain the aligned dataset $\hat{\mathcal{X}_{i}}=\left\{\hat{\mathcal{X}}_{i,1},\hat{\mathcal{X}}_{i,2},\cdots,\hat{\mathcal{X}}_{i,K}\right\}$.

        So far, we have introduced the procedure of stain alignment (i.e., $\mathcal{A}$).
        After all clients have applied $\mathcal{A}$ to their datasets, a tumor classifier can be trained in the FL system (i.e., the third step in Fig.~\ref{fig:workflow}).

\section{Experiments}
    \subsection{Experimental Settings}
\begin{table}[ht]
    \centering\setlength{\tabcolsep}{3.5pt}
    \begin{tabular}{l|ccc|c}
        \toprule
        Building Block&Hidden size&\#Heads&\#Params&FD\\
        \midrule
        MLP&32&--&4.7K&18.91\\
        Transformer&32&8&13.2K&0.40\\
        \bottomrule
    \end{tabular}
    \caption{Configuration details of diffusion models with different types of building blocks, and comparison of their FD scores. Note that both models have a single layer.}\label{tab:configuration}
\end{table}
        \subsubsection{Datasets}
            We evaluated our proposed method on three datasets, including Mitos \& Atypia 14 (MA14)~\cite{roux2014detection}, CAMELYON17 (C17)~\cite{8447230c17}, and AGGC22 (A22)~\cite{huo2024comprehensiveaggc}.
            Each dataset contains whole-slide images (WSIs) stained with hematoxylin and eosin (H{\&}E) dyes, so the number of stains is 2.
            Since we focus on patch-level classification in this work, we followed~\cite{guo2019fast} to extract histopathological image patches from each WSI.
            For more detailed information on each dataset, please refer to Table~\ref{tab:ma14_numbers},~\ref{tab:c17_numbers}, and~\ref{tab:a22_numbers} (in the Appendix).
            In addition, we provide a few sample images of each dataset in Figs.~\ref{fig:ma14_dataset},~\ref{fig:c17_dataset}, and~\ref{fig:a22_dataset} (in the Appendix).
        \subsubsection{Baselines}
            We adopted FedAvg~\cite{mcmahan2017communication}, FedProx~\cite{li2020federated}, FedSAM~\cite{qu2022generalized}, FedLESAM~\cite{pmlr-v235-fan24c}, and FedFA~\cite{zhou2023fedfa} as baselines to evaluate performance before and after applying FedSDA.
            Additionally, we compared the performance with CCST~\cite{chen2023federated} and the amplitude normalization in HarmoFL~\cite{jiang2022harmofl} (denoted by amp-norm), since these methods address non-IID data issues solely from the data distribution perspective, similar to FedSDA.
            For a clearer understanding of amp-norm, please refer to the pseudocode in Algorithm~\ref{alg:an-harmofl} (in the Appendix).
        \subsubsection{Evaluation Metrics}
            Due to the class-imbalanced nature of the A22 dataset, we evaluated tumor classification performance using the area under the receiver operating characteristic curve (AUROC)~\cite{AUROC} and the area under the precision-recall curve (AUPRC)~\cite{AUPRC}.
            For the class-balanced C17 dataset, we used AUROC only.
            On the other hand, following~\cite{tsai2025test,zingman2024comparative}, we adopted the Structural Similarity Index (SSIM)~\cite{wang2004image}, Wasserstein Distance (WD)~\cite{ramdas2017wasserstein}, Fr\'{e}chet Inception Distance (FID)~\cite{heusel2017gans}, and Kernel Inception Distance (KID)~\cite{binkowski2018demystifying} to assess image quality.
            SSIM measured the loss of structural information, which is important for tumor classification.
            WD computed the discrepancy between two images in terms of chromatic appearance and was adopted because stains are a pivotal factor in color variations in histopathological images.
            Both FID and KID~(an improved version of FID) were adopted to measure the distance between two data distributions.
        \subsubsection{Implementation Details}
            Each domain in a dataset is considered a client, where each domain refers to a subset with similar stains from the same hospital.
            We trained the diffusion model used in FedSDA with 3 communication rounds, 300 local epochs, a batch size of 65,536, and AdamW optimizer~\cite{loshchilov2017decoupled} with a weight decay of $3 e^{-2}$ and a learning rate of $2 e^{-4}$.
            We provide the configuration of the single-layer transformer model that we used in Table~\ref{tab:configuration}.            
            For tumor classification, we adopted DenseNet-121~\cite{huang2017densely} as the classifier due to its wide adoption for histopathological images.
            We trained classifiers on the C17 and A22 datasets with 100 communication rounds, one local epoch, a batch size of 128, cross-entropy loss, and SGD optimizer with a learning rate of $1 e^{-3}$.
            We implemented all the methods using PyTorch and conducted the experiments on an NVIDIA Tesla V100 GPU.
\begin{table}[ht]
    \centering\setlength{\tabcolsep}{7.5pt}
    \begin{tabular}{l|c|ccc}
        \toprule
        \multicolumn{1}{c|}{\multirow{2}{*}{Method}}&C17&\multicolumn{2}{c}{A22}\\
        \cmidrule{2-4}
        &AUROC↑&AUROC↑&AUPRC↑\\
        \midrule\midrule
        FedAvg&90.97&78.99&46.60\\
        \ \ w/ FedSDA&$\textbf{93.84}$&\textbf{90.38}&\textbf{64.20}\\
        \midrule
        FedProx&90.80&79.82&45.71\\
        \ \ w/ FedSDA&\textbf{93.89}&\textbf{90.44}&\textbf{64.89}\\
        \midrule
        FedSAM&90.87&82.17&48.21\\
        \ \ w/ FedSDA&\textbf{93.04}&\textbf{92.09}&\textbf{67.45}\\
        \midrule
        FedLESAM&90.17&79.42&44.39\\
        \ \ w/ FedSDA&\textbf{92.68}&\textbf{90.20}&\textbf{64.34}\\
        \midrule\midrule
        FedFA&92.31&82.62&48.45\\
        \ \ w/ FedSDA&\textbf{94.92}&\textbf{91.79}&\textbf{66.86}\\
        \bottomrule
    \end{tabular}\caption{Results of tumor classification before and after applying FedSDA.}\label{tab:tumor_classification_main}
\end{table}
\begin{figure}[b]
    \centering\includegraphics[width=\linewidth]{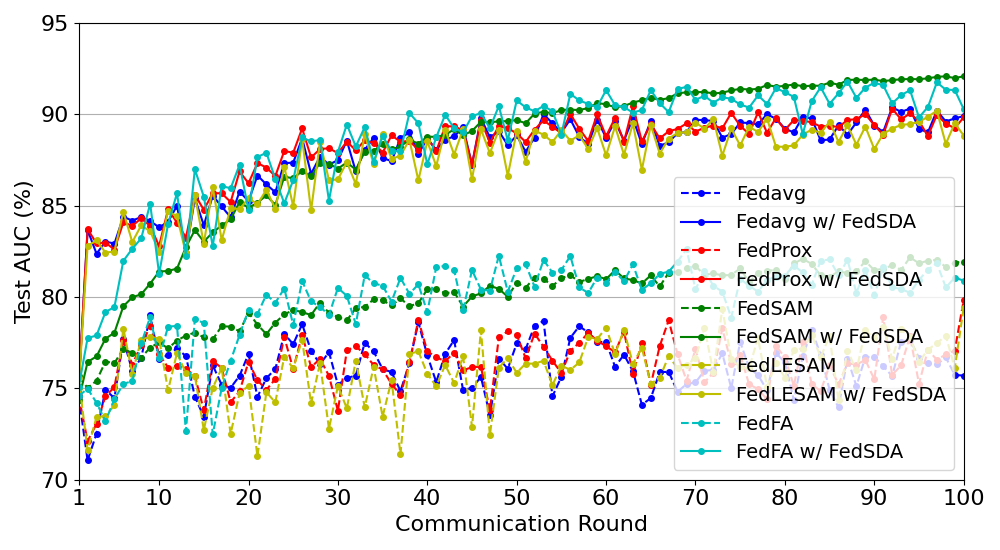}\caption{Test AUC vs. communication rounds on the A22 dataset before and after applying FedSDA.}\label{fig:a22_testauc_communicationround}
\end{figure}
\begin{table*}[ht]
    \centering\setlength{\tabcolsep}{3.5pt}
    \begin{tabular}{l|c|cc|cccc}
        \toprule
        \multicolumn{1}{c|}{\multirow{2}{*}{Method}}&C17&\multicolumn{2}{c|}{A22}&\multicolumn{4}{|c}{MA14 \& C17 \& A22}\\
        \cmidrule{2-8}
        &AUROC↑&AUROC↑&AUPRC↑&SSIM↑&FID↓&KID↓&WD↓\\
        \midrule\midrule
        FedAvg&90.97&78.99&46.60&1.0000&55.092&0.1986&0.0013\\
        \ w/ CCST&71.89&80.09&38.43&0.2462&\textbf{20.777}&\textbf{0.0599}&\textbf{0.0007}\\
        \ w/ amp-norm&77.49&77.01&42.83&0.9731&54.774&0.1956&0.0012\\
        \ w/ FedSDA&\textbf{93.84}&\textbf{90.38}&\textbf{64.20}&\textbf{0.9969}&40.311&0.1365&0.0009\\
        \bottomrule
    \end{tabular}\caption{Comparison of tumor classification and image quality (averaged over three datasets, i.e., MA14, C17, and A22) across different modules embedded in FedAvg.}\label{tab:tumor_classification_main_I}
\end{table*}
    \subsection{Main Results}
        \subsubsection{Evaluation of Tumor Classification}
            First, we applied FedSDA to the five baselines (i.e.,~FedAvg~\cite{mcmahan2017communication}, FedProx~\cite{li2020federated}, FedSAM~\cite{qu2022generalized}, FedLESAM~\cite{pmlr-v235-fan24c}, and FedFA~\cite{zhou2023fedfa}) to verify whether the performance can be improved with the use of FedSDA.
            Table~\ref{tab:tumor_classification_main} demonstrates that incorporating FedSDA yields a significant performance improvement across the baselines.
            Note that FedAvg combined with FedSDA can outperform the other four baselines by a significant margin.
            Additionally, Fig.~\ref{fig:a22_testauc_communicationround} depicts the convergence curves corresponding to the results in Table~\ref{tab:tumor_classification_main} and demonstrates that incorporating FedSDA enables convergence not only to a more optimal solution but also at a faster rate.
            Second, the results of different modules embedded in FedAvg~\cite{mcmahan2017communication} are compared in Table~\ref{tab:tumor_classification_main_I}, and it can be observed that FedSDA significantly outperforms the other modules, regardless of whether the dataset is imbalanced.
            
        \subsubsection{Evaluation of Image Quality}
\begin{figure}[b]
    \centering\includegraphics[width=\linewidth]{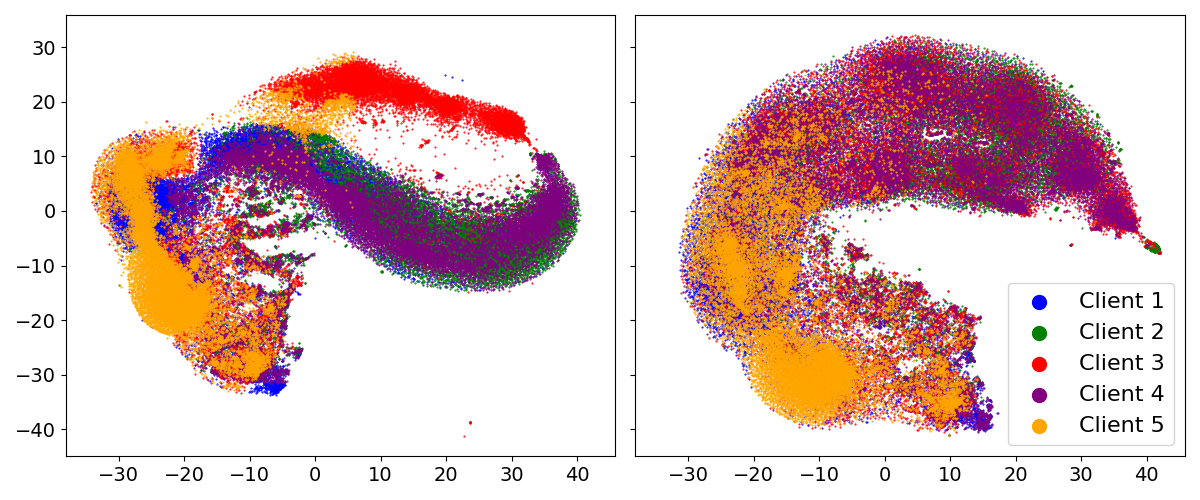}\caption{Visualizations of the C17 dataset before and after applying FedSDA by t-SNE. Please refer to Figs.~\ref{fig:ma14_visualization},~\ref{fig:c17_visualization}, and~\ref{fig:a22_visualization} (in the Appendix) for detailed results on all datasets.}\label{fig:c17_ba_tsne}
\end{figure}

            Table~\ref{tab:tumor_classification_main_I} also shows a comparison of image quality across different embedded modules.
            It can be observed that FedSDA outperforms both the baseline without FedSDA and the one using amp-norm.
            We also observe that CCST~\cite{chen2023federated} outperforms FedSDA in terms of FID, KID, and WD; however, it loses significantly more structural information, as indicated by a drop in SSIM to 0.2462, which substantially compromises its effectiveness for tumor classification.
            This is clearly evidenced in Table~\ref{tab:tumor_classification_main_I}.
            Finally, we visualize histopathological images before and after applying FedSDA in Fig.~\ref{fig:c17_ba_tsne}.
            After applying FedSDA, the low-dimensional features of each client's data tend to be distributed more similarly than before.
            Additional visualization results are provided in the Appendix.
            Specifically, we include visualizations of the stain matrices (Figs.~\ref{fig:ma14_stain_matrices},~\ref{fig:c17_stain_matrices}, and~\ref{fig:a22_stain_matrices} in the Appendix), comprising both those decomposed from the histopathological images of each client and those generated by the diffusion model.
\begin{figure}[ht]
    \centering\includegraphics[width=\linewidth]{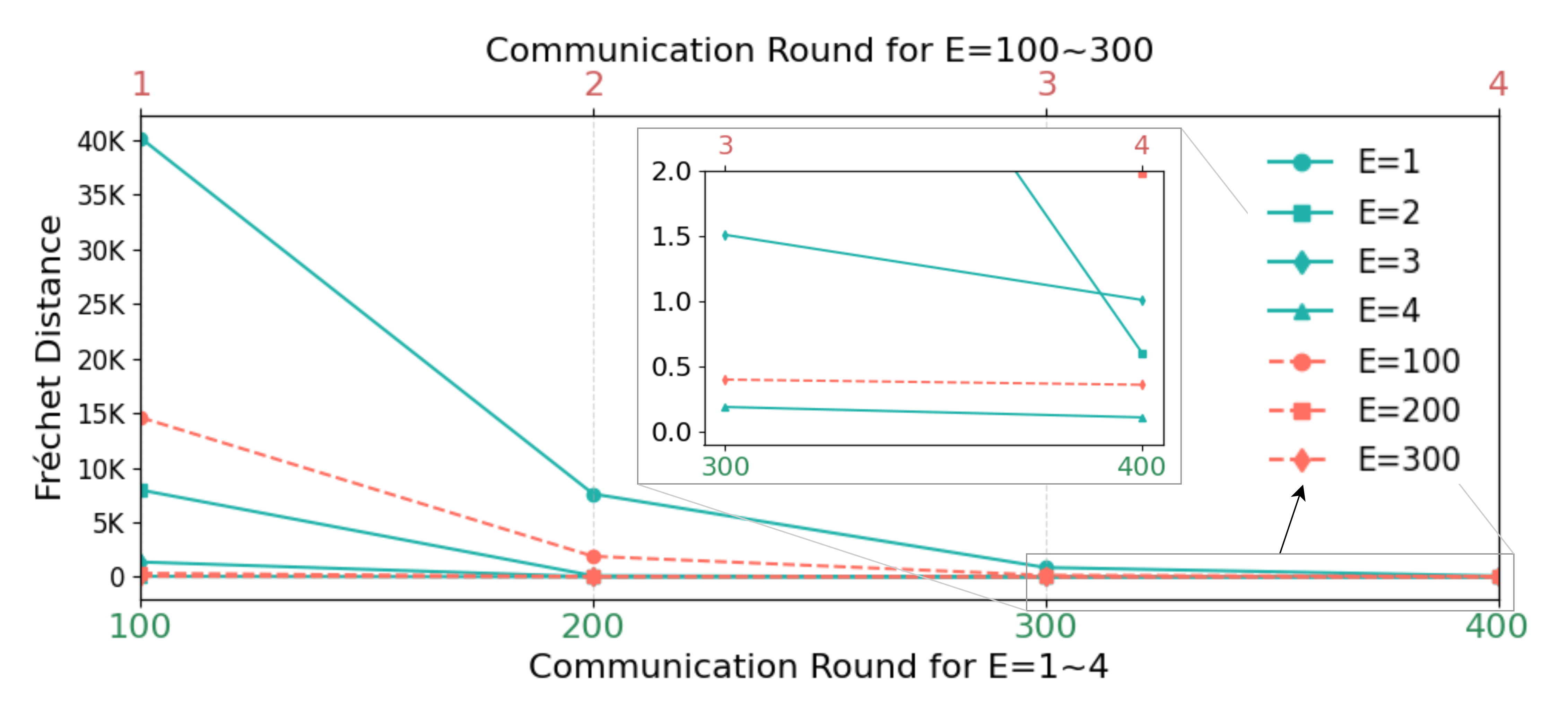}\caption{Fr\`{e}chet distance vs. communication rounds under different numbers of local epochs on the C17 dataset.}\label{fig:ablation_study_fd}
\end{figure}
    \subsection{Ablation Study on Diffusion Model in FedSDA}
        The ablation studies conducted here are measured and compared in terms of the Fr\'{e}chet Distance (FD) between two sets of stain matrices.
        \subsubsection{Communication Rounds vs. Local Epochs}
            To examine the effect of the number of local epochs on communication rounds, we set the number of local epochs as 1, 2, 3, 4, 100, 200, and 300, and recorded the corresponding number of communication rounds required for convergence. 
            As shown in Fig.~\ref{fig:ablation_study_fd}, many more communication rounds are required for convergence when using fewer local epochs (i.e., 1 -- 4).
            However, training converges with an FD of 0.40 and a WD of 0.07 in only 3 communication rounds when the number of local epochs was set to 300.
            This observation suggests that increasing the number of local epochs can significantly reduce communication costs while maintaining model performance.
        \subsubsection{Network Architecture}
            With the number of communication rounds and local epochs set to 3 and 300, respectively, we compared the performance of a single-layer transformer model with a single-layer MLP model.
            As shown in Table~\ref{tab:configuration}, a much more lightweight network architecture with MLP as the main building block significantly deteriorates performance.

\section{Conclusion}
    In this paper, we propose FedSDA to address the challenge of non-IID histopathological images with feature distribution shifts in a federated environment.
    By leveraging stain separation and fitting a target distribution using a diffusion model without directly training on raw data, FedSDA mitigates stain distribution shifts across clients while preventing raw data from being directly exposed to the risks of leakage in a federated environment.
    Extensive experiments show that FedSDA not only improves the performance of optimization-based methods but also outperforms distribution-based methods.
    This work highlights the potential of focusing on data-centric solutions within FL, particularly for applications involving complex, non-IID medical image datasets.
    
\section{Acknowledgments}
This work was supported by the National Science and Technology
Council (NSTC), Taiwan, ROC, under Grants NSTC 113-2634-F-006-002 and 114-2221-E-001 -010 -MY2.

\bibliography{aaai}
\clearpage

\appendix
\section{Pseudocode for FedSDA}
    In this section, we present two pseudocodes, including
    \begin{itemize}
        \item Target distribution fitting in Algorithm~\ref{alg:train_cdm}
        \item Stain alignment in Algorithm~\ref{alg:stain_alignment}.
    \end{itemize}

    \begin{algorithm}
        \caption{Target Distribution Fitting}\label{alg:train_cdm}
        \begin{algorithmic}
            \STATE \textbf{Input:} Number of client $K$, number of communication rounds $R$, number of local epochs $E$, local batch size $B$, learning rate $\eta$, number of diffusion timesteps $T$, variance schedule $\left\{\beta_{1},\beta_{2},\ldots,\beta_{T}\right\}$, local histopathological image datasets $\left\{\mathcal{X}_{1},\mathcal{X}_{2},\ldots,\mathcal{X}_{K}\right\}$, initialized $\mathcal{G}_{\theta}$ (parameterized by $\theta$)
            \STATE \textbf{Output:} Conditional diffusion model $\mathcal{G}_{\theta}$
            \\[-2.5mm] \hrulefill
            \STATE $\left|\mathcal{X}\right|\leftarrow\sum^{K}_{k=1}\left|\mathcal{X}_{k}\right|$
            \STATE $\theta_{0}\leftarrow\theta$
            \STATE \textit{\# Obtaining stain matrices using stain separation}
            \FOR{$k=1$ \TO $K$}
            \STATE $\mathcal{W}_{k}\leftarrow\left\{\right\}$
            \FOR{$\mathbf{x}\in\mathcal{X}_{k}$}
            \STATE $\mathbf{w},\underline{\quad}\leftarrow\mathcal{S}(\mathbf{x})$\COMMENT{Eq.~(\ref{eq:SR})}
            \STATE $\mathcal{W}_{k}\leftarrow\mathcal{W}_{k}\bigcup\left\{\mathbf{w}\right\}$
            \ENDFOR
            \ENDFOR
            \STATE \textit{\# Training a conditional diffusion model}
            \FOR{$r=1$ \TO $R$}
            \STATE \textit{\# Local training}
            \FOR{$k=1$ \TO $K$}
            \STATE $\theta^{k}_{r}\leftarrow\theta_{r-1}$
            \STATE $\mathcal{B}_{k}\leftarrow$ (split $\mathcal{W}_{k}$ into batches of size $B$)
            \FOR{$e=1$ \TO $E$}
            \FOR{$b\in\mathcal{B}_{k}$}
            \STATE $\mathcal{L}\leftarrow{0}$
            \FOR{$\mathbf{w}\in{b}$}
            \STATE $t\sim\mbox{Uniform}(\left\{1,2,\ldots,T\right\})$
            \STATE $\mathbf{\epsilon}\sim\mathcal{N}(\mathbf{0},\mathbf{I})$
            \STATE $\bar{\alpha}_{t}\leftarrow\prod^{t}_{s=1}(1-\beta_{s})$
            \STATE $\hat{\mathcal{L}}\leftarrow\left\|\epsilon-\mathcal{G}_{\theta^{k}_{r}}(\sqrt{\bar{\alpha}_{t}}\mathbf{w}+\sqrt{1-\bar{\alpha}_{t}}\mathbf{\epsilon},t,k)\right\|^{2}_{2}$
            \STATE $\mathcal{L}\leftarrow\mathcal{L}+\hat{\mathcal{L}}$
            \ENDFOR
            \STATE $\theta^{k}_{r}\leftarrow\theta^{k}_{r}-\eta\nabla_{\theta^{k}_{r}}\frac{\mathcal{L}}{\left|b\right|}$
            \ENDFOR
            \ENDFOR
            \ENDFOR
            \STATE \textit{\# Central server aggregation}
            \STATE $\theta_{r}\leftarrow\sum^{K}_{k=1}\frac{\left|\mathcal{X}_{k}\right|}{\left|\mathcal{X}\right|}\theta^{k}_{r}$
            \ENDFOR
            \STATE $\theta\leftarrow\theta_{R}$
            \RETURN $\mathcal{G}_{\theta}$
        \end{algorithmic}
    \end{algorithm}

    \begin{algorithm}
        \caption{Stain Alignment}\label{alg:stain_alignment}
        \begin{algorithmic}
            \STATE \textbf{Input:} Number of client $K$, local histopathological image datasets $\left\{\mathcal{X}_{1},\mathcal{X}_{2},\ldots,\mathcal{X}_{K}\right\}$, conditional diffusion model $\mathcal{G}$
            \STATE \textbf{Output:} Aligned datasets $\hat{\mathcal{X}}_{1},\hat{\mathcal{X}}_{2},\ldots,\hat{\mathcal{X}}_{K}$
            \\[-2.5mm] \hrulefill
            \FOR{$k=1$ \TO $K$}
            \STATE $n\leftarrow\left|\mathcal{X}_{k}\right|/{K}$
            \STATE $\left\{\mathcal{X}_{k,i}\right\}^{K}_{i=1}\leftarrow$ (split $\mathcal{X}_{k}$ into $K$ subsets of size $n$)
            \FOR{$i=1$ \TO $K$}
            \STATE $\hat{\mathcal{X}}_{k,i}\leftarrow\left\{\right\}$
            \FOR{$\mathbf{x}\in\mathcal{X}_{k,i}$}
            \STATE $\underline{\quad},\mathbf{h}\leftarrow\mathcal{S}(\mathbf{x})$\COMMENT{Eq.~(\ref{eq:SR})}
            \STATE $\mathbf{w}\leftarrow\mathcal{G}(i)$\COMMENT{Sampling a stain matrix}
            \STATE $\hat{\mathbf{x}}\leftarrow\mathcal{R}(\mathbf{w},\mathbf{h})$\COMMENT{Eq.~(\ref{eq:SR})}
            \STATE $\hat{\mathcal{X}}_{k,i}\leftarrow\hat{\mathcal{X}}_{k,i}\bigcup\left\{\hat{\mathbf{x}}\right\}$
            \ENDFOR
            \ENDFOR
            \STATE $\hat{\mathcal{X}}_{k}\leftarrow\bigcup^{K}_{i=1}\hat{\mathcal{X}}_{k,i}$
            \ENDFOR
            \RETURN $\hat{\mathcal{X}}_{1},\hat{\mathcal{X}}_{2},\ldots,\hat{\mathcal{X}}_{K}$
        \end{algorithmic}
    \end{algorithm}

\section{Pseudocode for Amplitude Normalization in HarmoFL}
    In this section, we present the pseudocode for amplitude normalization in HarmoFL~\cite{jiang2022harmofl} (i.e., amp-norm) in Algorithm~\ref{alg:an-harmofl}.

    \begin{algorithm}
        \caption{amp-norm}\label{alg:an-harmofl}
        \begin{algorithmic}
            \STATE \textbf{Input:} Number of client $K$, batch size $B$, decay factor $v$, local histopathological image datasets $\left\{\mathcal{X}_{1},\mathcal{X}_{2},\ldots,\mathcal{X}_{K}\right\}$
            \STATE \textbf{Output:} Amplitude normalized datasets $\hat{\mathcal{X}}_{1},\hat{\mathcal{X}}_{2},\ldots,\hat{\mathcal{X}}_{K}$
            \\[-2.5mm] \hrulefill
            \STATE \textit{\# Computing the average amplitude}
            \FOR{$k=1$ \TO $K$}
            \STATE $\mathcal{A}_{k}\leftarrow\mathbf{0}$
            \STATE $\mathcal{B}_{k}\leftarrow$ (split $\mathcal{X}_{k}$ into batches of size $B$)
            \FOR{$b\in\mathcal{B}_{k}$}
            \STATE $\hat{\mathcal{A}}_{k}\leftarrow\mathbf{0}$
            \FOR{$\mathbf{x}\in{b}$}
            \STATE \textit{\# Let $\mathcal{A}_{\mathbf{x}},\mathcal{P}_{\mathbf{x}}=\mathcal{F}(\mathbf{x})$,}
            \STATE \textit{\# where $\mathcal{F}$ is the Fourier transform,}
            \STATE \textit{\# $\mathcal{A}_{\mathbf{x}}$ is the amplitude for $\mathbf{x}$,}
            \STATE \textit{\# and $\mathcal{P}_{\mathbf{x}}$ is the phase for $\mathbf{x}$.}
            \STATE $\mathcal{A},\underline{\quad}\leftarrow\mathcal{F}(\mathbf{x})$
            \STATE $\hat{\mathcal{A}}_{k}\leftarrow\hat{\mathcal{A}}_{k}+\mathcal{A}$
            \ENDFOR
            \STATE $\mathcal{A}_{k}\leftarrow(1-v)\mathcal{A}_{k}+\frac{v}{B}\hat{\mathcal{A}}_{k}$
            \ENDFOR
            \ENDFOR
            \STATE $\overline{\mathcal{A}}\leftarrow\frac{1}{K}\sum^{K}_{k=1}\mathcal{A}_{k}$
            \STATE \textit{\# Normalizing amplitudes of histopathological images}
            \FOR{$k=1$ \TO $K$}
            \STATE $\hat{\mathcal{X}}_{k}\leftarrow\left\{\right\}$
            \FOR{$\mbf{x}\in\mathcal{X}_{k}$}
            \STATE \textit{\# $\mathcal{F}^{-1}$ is the inverse Fourier transform.}
            \STATE $\underline{\quad},\mathcal{P}\leftarrow\mathcal{F}(\mbf{x})$
            \STATE $\hat{\mbf{x}}\leftarrow\mathcal{F}^{-1}(\overline{\mathcal{A}},\mathcal{P})$
            \STATE $\hat{\mathcal{X}}_{k}\leftarrow\hat{\mathcal{X}}_{k}\bigcup\left\{\hat{\mbf{x}}\right\}$
            \ENDFOR
            \ENDFOR
            \RETURN $\hat{\mathcal{X}}_{1},\hat{\mathcal{X}}_{2},\ldots,\hat{\mathcal{X}}_{K}$
        \end{algorithmic}
    \end{algorithm}

\section{Dataset Description}
    \subsection{Mitos \& Atypia 14}
\begin{figure*}[ht]
            \centering\includegraphics[width=1\linewidth]{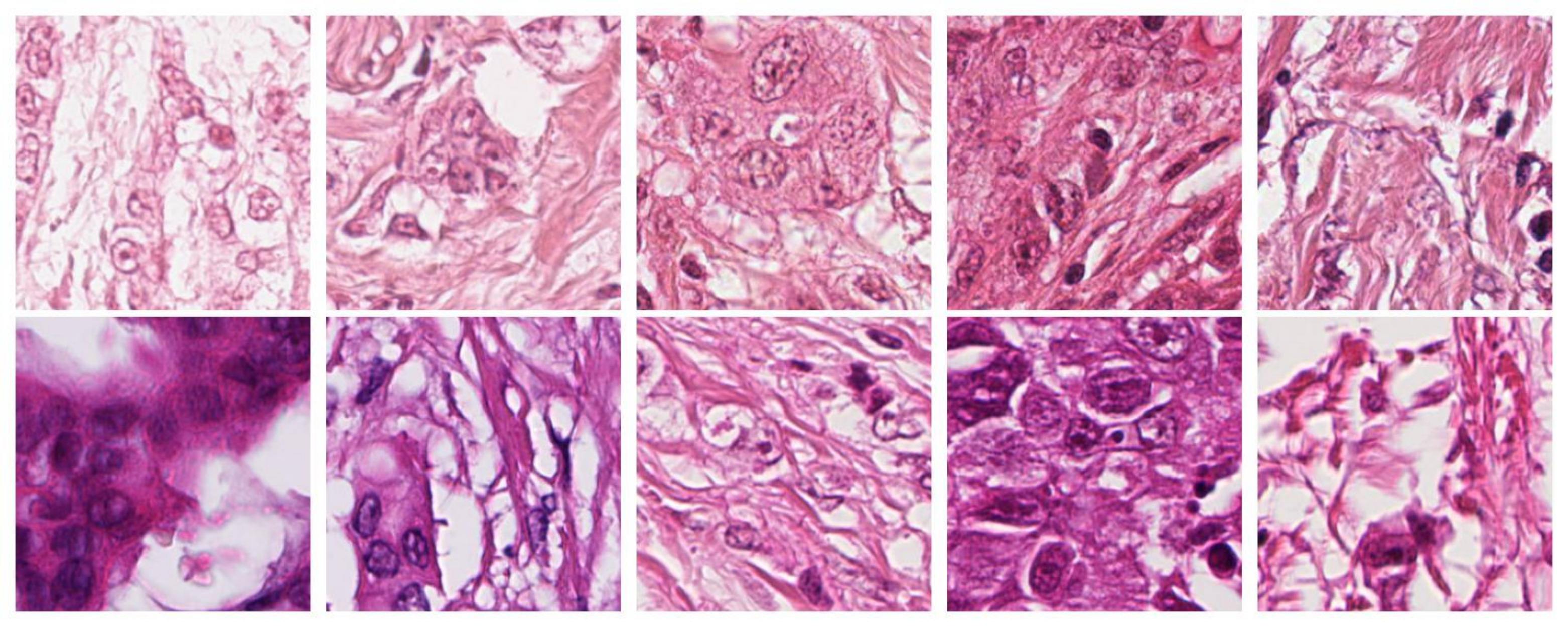}
            \caption{Sample images of the Mitos \& Atypia 14~\cite{roux2014detection} dataset. Top row: Client 1. Bottom row: Client 2.}
            \label{fig:ma14_dataset}
        \end{figure*}

        Mitos \& Atypia 14 (i.e., MA14) contains the whole-slide images (WSIs) of hematoxylin and eosin (H{\&}E) stained breast cancer tissue.
        Since the WSIs are produced from two different scanners, they are regarded as being originated from two different clients accordingly, thereby the number of clients in a federated learning environment is 2.
        The WSIs were split into histopathological patch images of size 256$\times$256 at 40$\times$ magnification.
        We provide the numbers of WSIs and histopathological patch images in each client in Table~\ref{tab:ma14_numbers} and present the sample images of each client in Fig.~\ref{fig:ma14_dataset}.
        \begin{table}[ht]
            \centering\setlength{\tabcolsep}{2.5pt}
            \begin{tabular}{c|cc}
                \toprule
                Client & 1 & 2\\
                \midrule\midrule
                WSIs & 1,136 & 1,136\\
                \midrule
                Histopathological Patch Images & 35,996 & 35,993\\
                \bottomrule
            \end{tabular}
            \caption{Numbers of WSIs and histopathological patches of each client in the Mitos \& Atypia 14 dataset.}
            \label{tab:ma14_numbers}
        \end{table}
    \subsection{CAMELYON17}
\begin{figure*}[ht]
            \centering\includegraphics[width=1\linewidth]{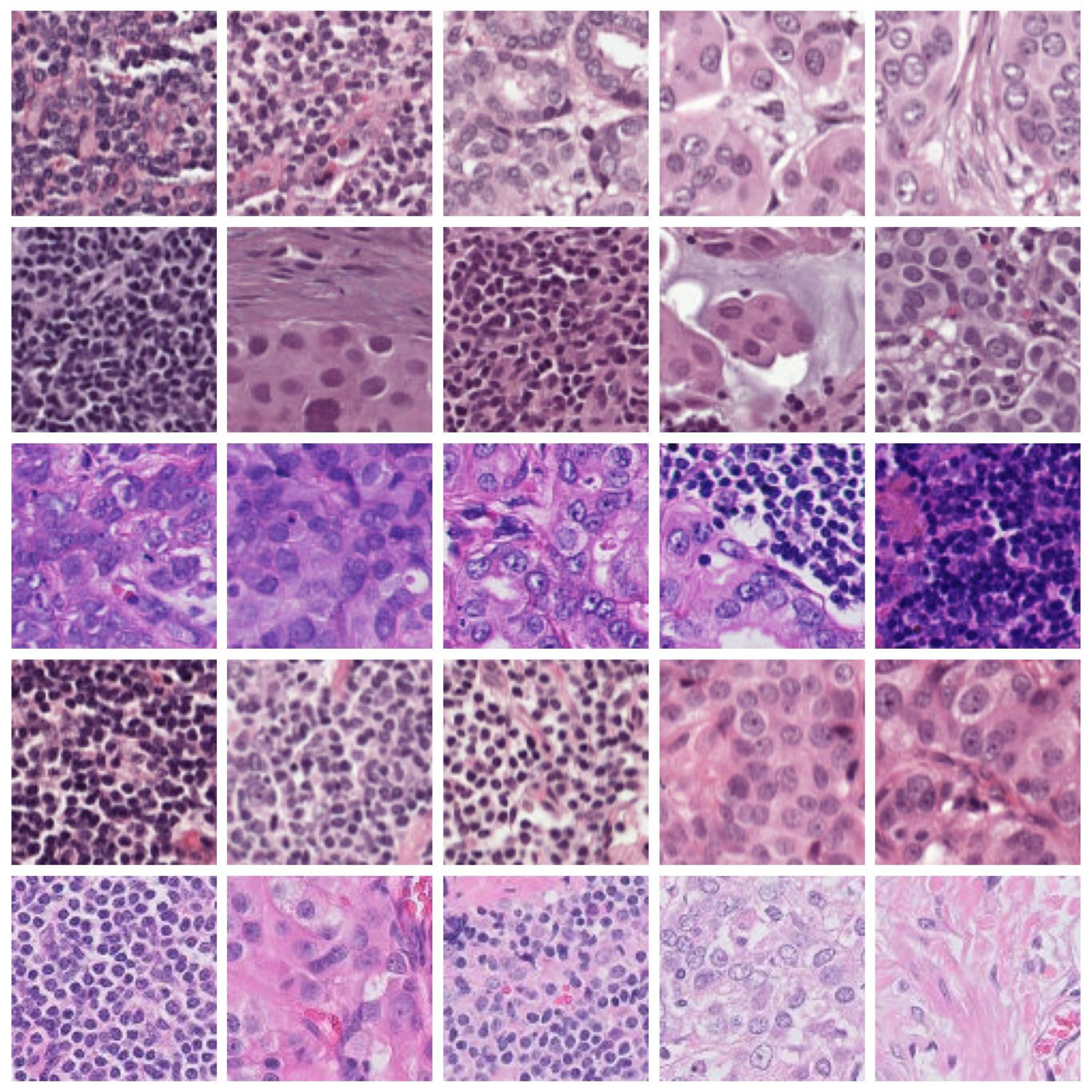}
            \caption{Sample images of the CAMELYON17 dataset~\cite{8447230c17}. Different rows, from top to bottom, indicate different clients.}
            \label{fig:c17_dataset}
        \end{figure*}

        CAMELYON17 (i.e., C17) contains the WSIs of H{\&}E stained lymph node sections.
        Since the WSIs were collected from 5 medical centers, the number of clients in federated learning is 5.
        The WSIs were split into histopathological patch images of size 96$\times$96 at 10$\times$ magnification.
        For data splitting in tumor classification, each client follows an 80/20 ratio as the training/test sets.
        We provide the numbers of WSIs and histopathological patches of each client in Table~\ref{tab:c17_numbers} and present the sample images of each client in Fig.~\ref{fig:c17_dataset}.
        \begin{table}[ht]
            \centering\setlength{\tabcolsep}{1.5pt}
            \begin{tabular}{c|ccccc}
                \toprule
                Client & 1 & 2 & 3 & 4 & 5\\
                \midrule\midrule
                WSIs & 10 & 10 & 10 & 10 & 10\\
                \midrule
                Histopathological & \multirow{2}{*}{59,042} & \multirow{2}{*}{34,832} & \multirow{2}{*}{85,053} & \multirow{2}{*}{129,140} & \multirow{2}{*}{146,721}\\
                patch Images &&&&&\\
                \bottomrule
            \end{tabular}
            \caption{Numbers of WSIs and histopathological patch images of each client in the CAMELYON17 dataset.}
            \label{tab:c17_numbers}
        \end{table}
    \subsection{AGGC22}
\begin{figure*}[ht]
            \centering\includegraphics[width=1\linewidth]{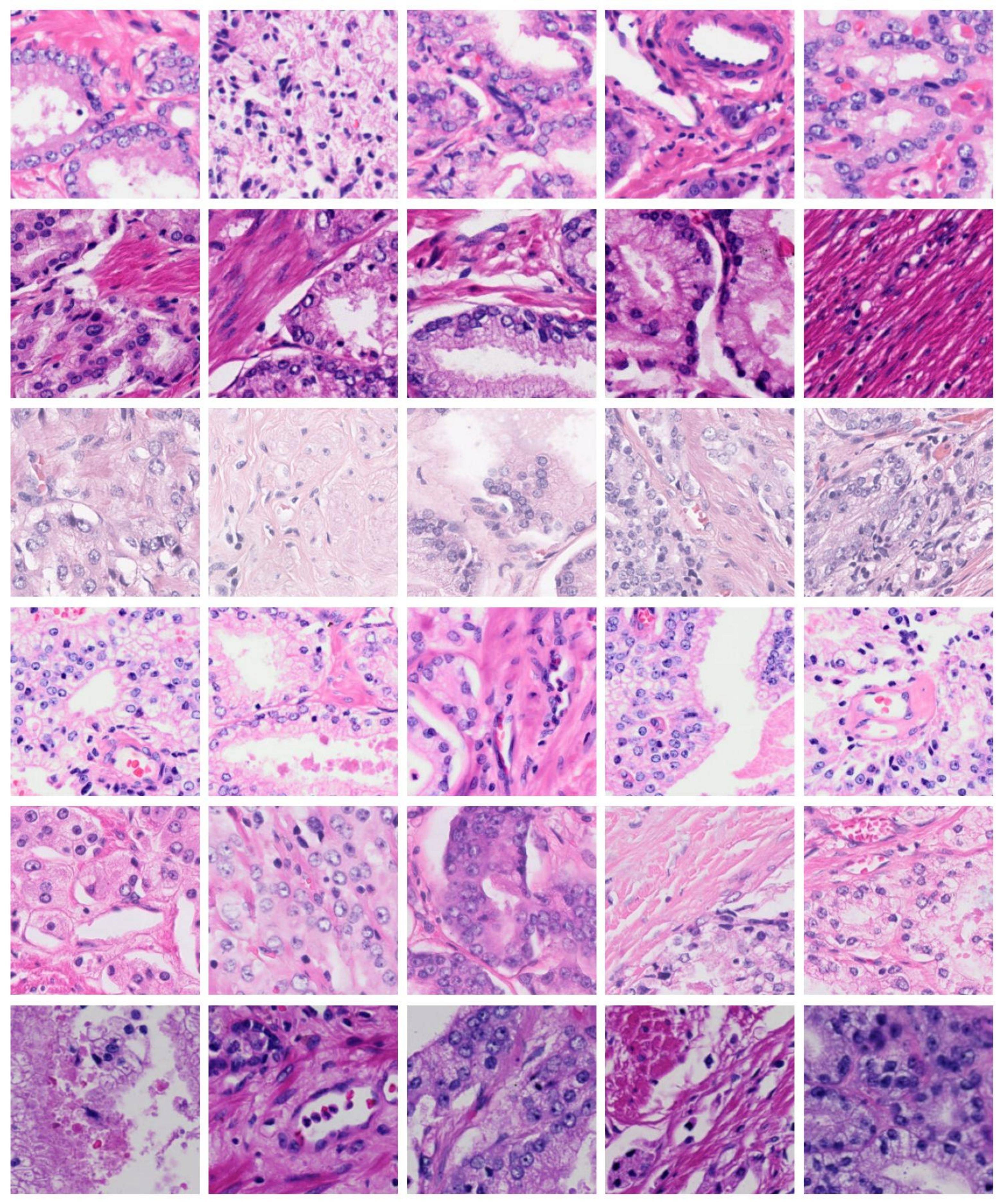}
            \caption{Sample images of the AGGC22 dataset~\cite{huo2024comprehensiveaggc}. Different rows, from top to bottom, indicate different clients.}
            \label{fig:a22_dataset}
        \end{figure*}

        AGGC22 (i.e., A22) contains H{\&}E stained WSIs of prostatectomy and biopsy specimens.
        Since each specimen is scanned by 6 scanners, the WSIs were classified into 6 clients based on which scanners they are scanned from.
        The WSIs were split into histopathological patch images of size 256$\times$256 at 20$\times$ magnification.
        For data splitting in tumor classification, each client follows a ratio of 60/40 as the training/test sets.
        We provide the numbers of WSIs and histopathological patch images in each client in Table~\ref{tab:a22_numbers} and present the sample images of each client in Fig.~\ref{fig:a22_dataset}.
        \begin{table}[ht]
            \centering\setlength{\tabcolsep}{1pt}
            \begin{tabular}{c|cccccc}
                \toprule
                Client & 1 & 2 & 3 & 4 & 5 & 6\\
                \midrule\midrule
                WSIs & 38 & 38 & 38 & 37 & 38 & 22\\
                \midrule
                Image & \multirow{2}{*}{119,860} & \multirow{2}{*}{254,366} & \multirow{2}{*}{210,151} & \multirow{2}{*}{152,151} & \multirow{2}{*}{217,518} & \multirow{2}{*}{115,163}\\
                Patches &&&&&\\
                \bottomrule
            \end{tabular}
            \caption{Numbers of WSIs and histopathological patch images of each client in the AGGC22 dataset.}
            \label{tab:a22_numbers}
        \end{table}
        
\section{Visualization Results}
\begin{figure*}[ht]
        \centering\includegraphics[width=1\linewidth]{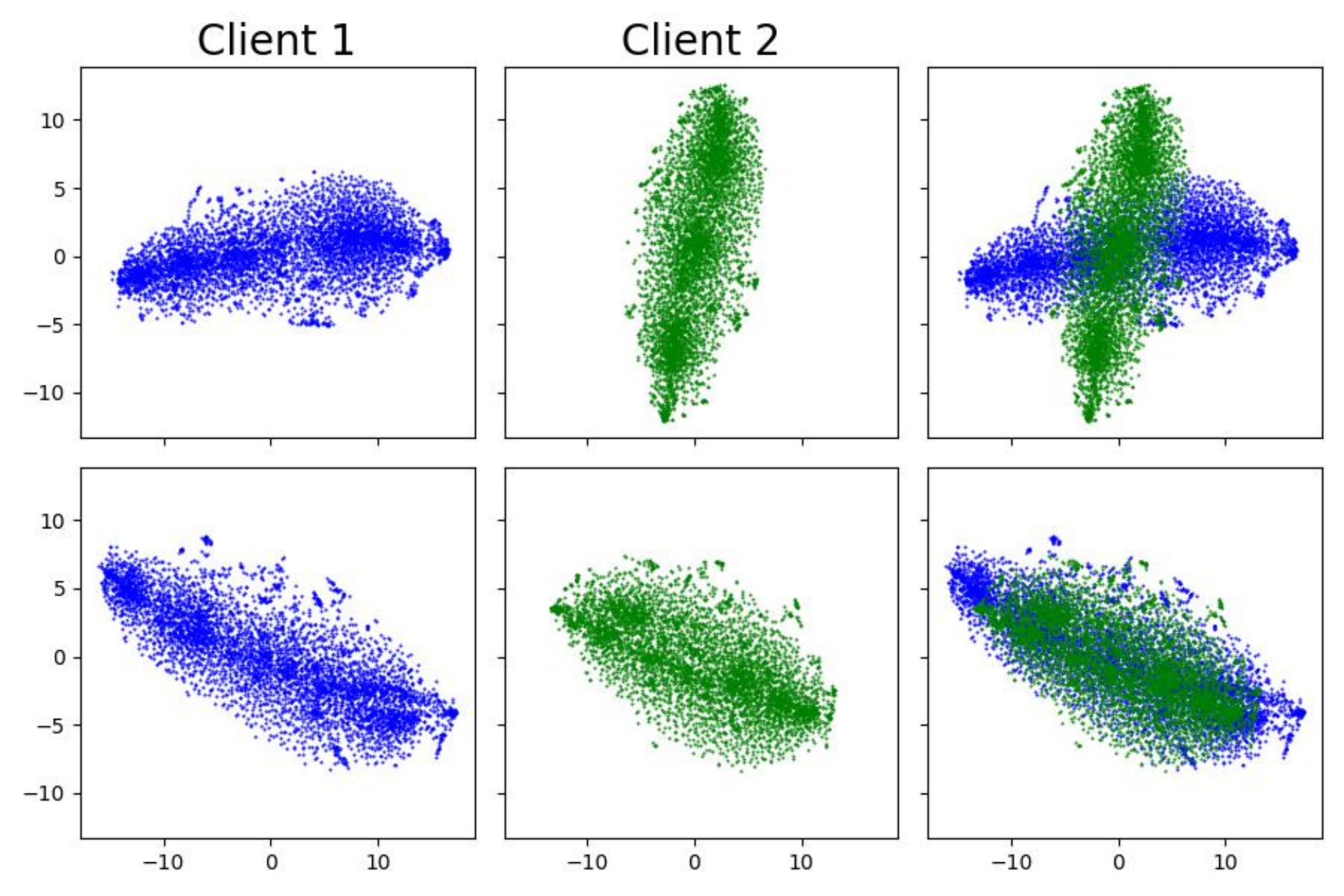}
        \caption{Visualizations of the MA$14$~\cite{roux2014detection} dataset for each client before and after applying FedSDA. Top row: Before applying FedSDA. Bottom row: After applying FedSDA. The rightmost column shows the results when those from the two clients are put together.}\label{fig:ma14_visualization}
    \end{figure*}

\begin{figure*}[ht]
        \centering\includegraphics[width=1\linewidth]{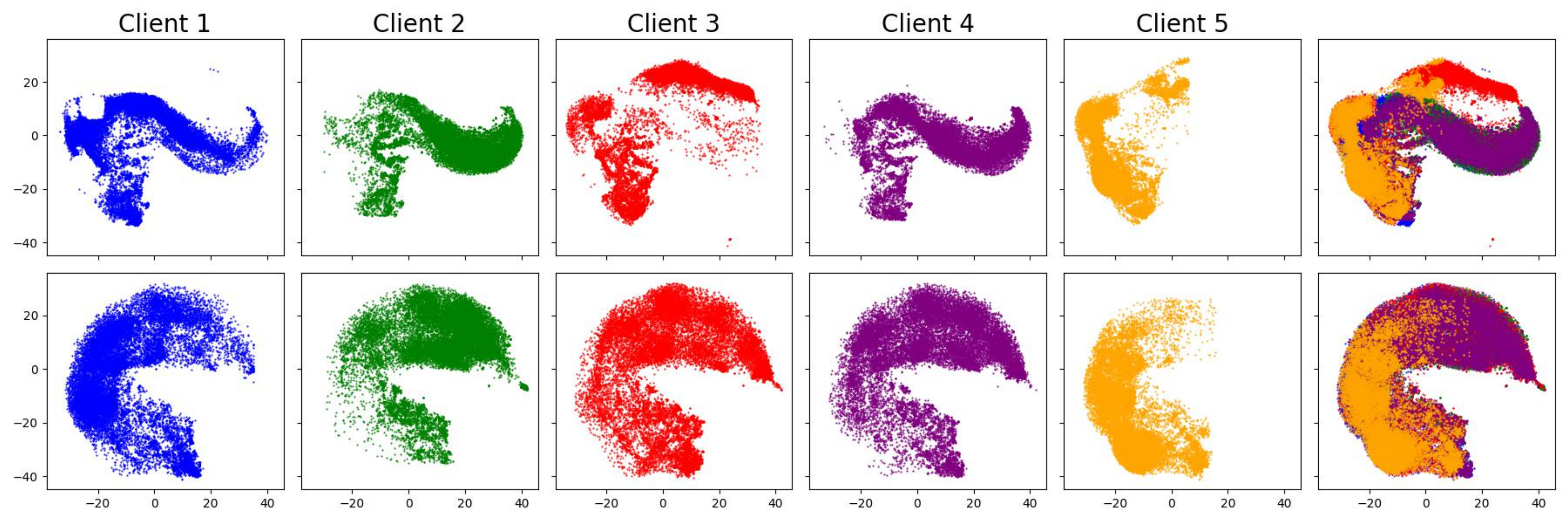}
        \caption{Visualizations of the C$17$ dataset~\cite{8447230c17} before and after applying FedSDA. Top row: Before applying FedSDA. Bottom row: After applying FedSDA. The rightmost column shows the results when those from the five clients are put together.}\label{fig:c17_visualization}
    \end{figure*}

\begin{figure*}[ht]
        \centering\includegraphics[width=1\linewidth]{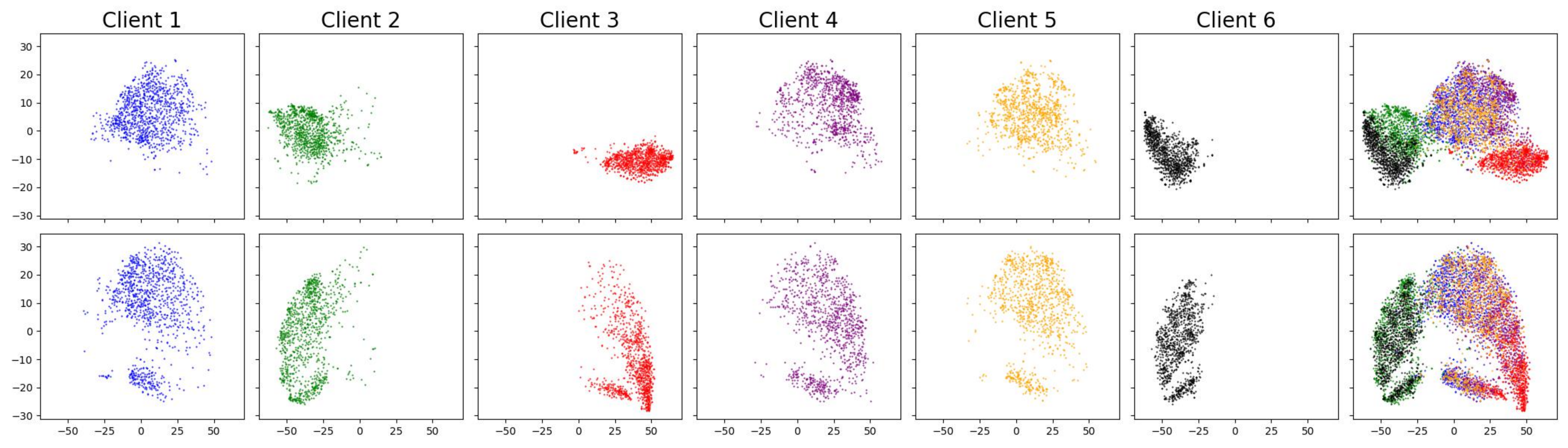}
        \caption{Visualizations of the  A22 dataset~\cite{huo2024comprehensiveaggc} before and after applying FedSDA. Top row: Before applying FedSDA. Bottom row: After applying FedSDA.  The rightmost column shows the results when those from the six clients are put together.}\label{fig:a22_visualization}
    \end{figure*}

    Here, we first present the t-SNE (t-distributed Stochastic Neighbor Embedding) visualizations of each dataset before and after applying FedSDA, as shown in Figs.~\ref{fig:ma14_visualization},~\ref{fig:c17_visualization}, and~\ref{fig:a22_visualization}.
    We observe that, after applying FedSDA, the low-dimensional features of each client's images tend to be distributed more identically compared to those without applying FedSDA.
    Note that although the low-dimensional features of each client's images in the A22 dataset are less identically distributed due to sampling bias, the results still show marginal improvement.

\begin{figure*}[ht]
        \centering\includegraphics[width=1\linewidth]{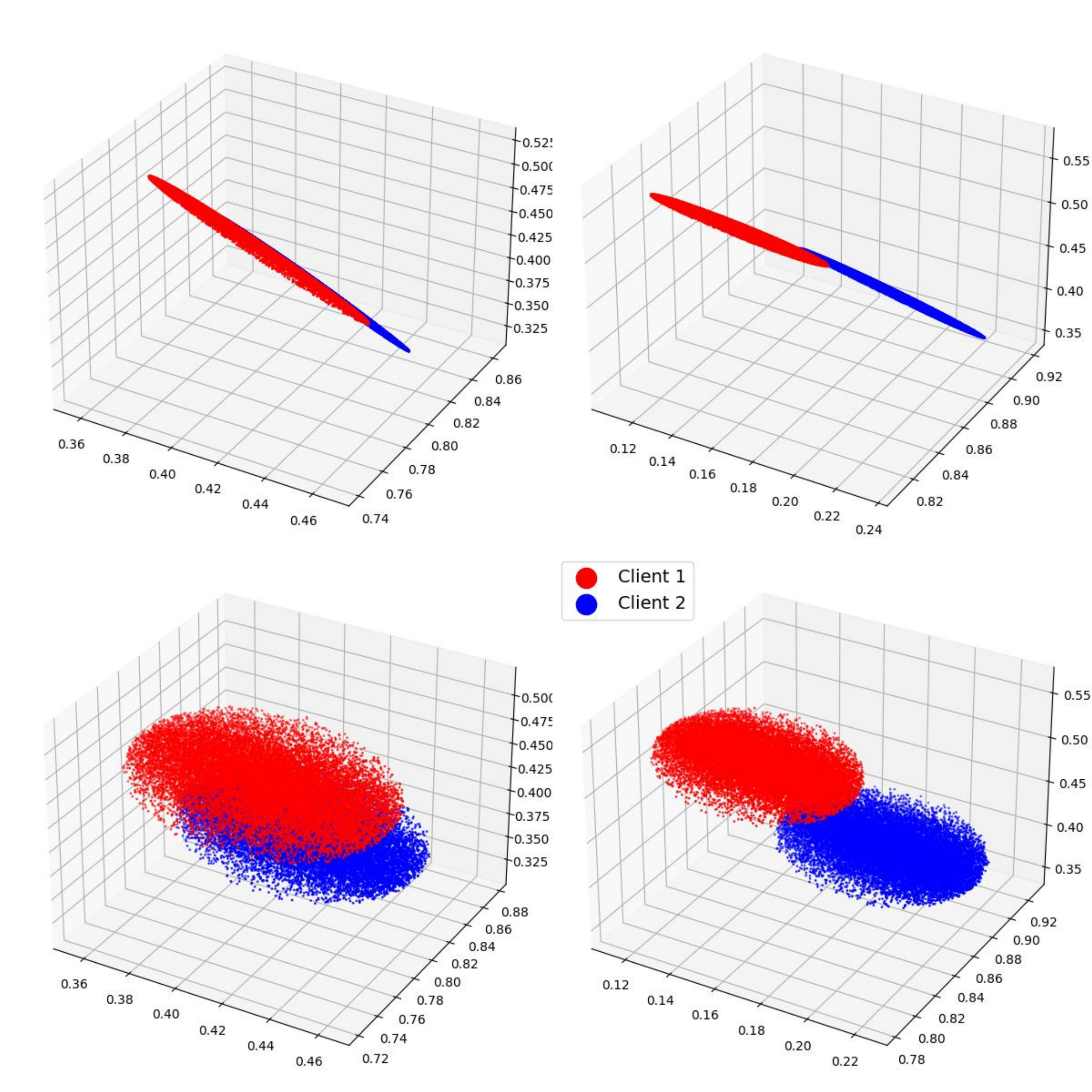}
        \caption{Stain matrix visualizations of the Mitos {\&} Atypia 14 dataset~\cite{roux2014detection}. Left column: H stain. Right column: E stain. Top row: Decomposition from histopathological images. Bottom row: Generated from a diffusion model.}\label{fig:ma14_stain_matrices}
    \end{figure*}

\begin{figure*}[ht]
        \centering\includegraphics[width=1\linewidth]{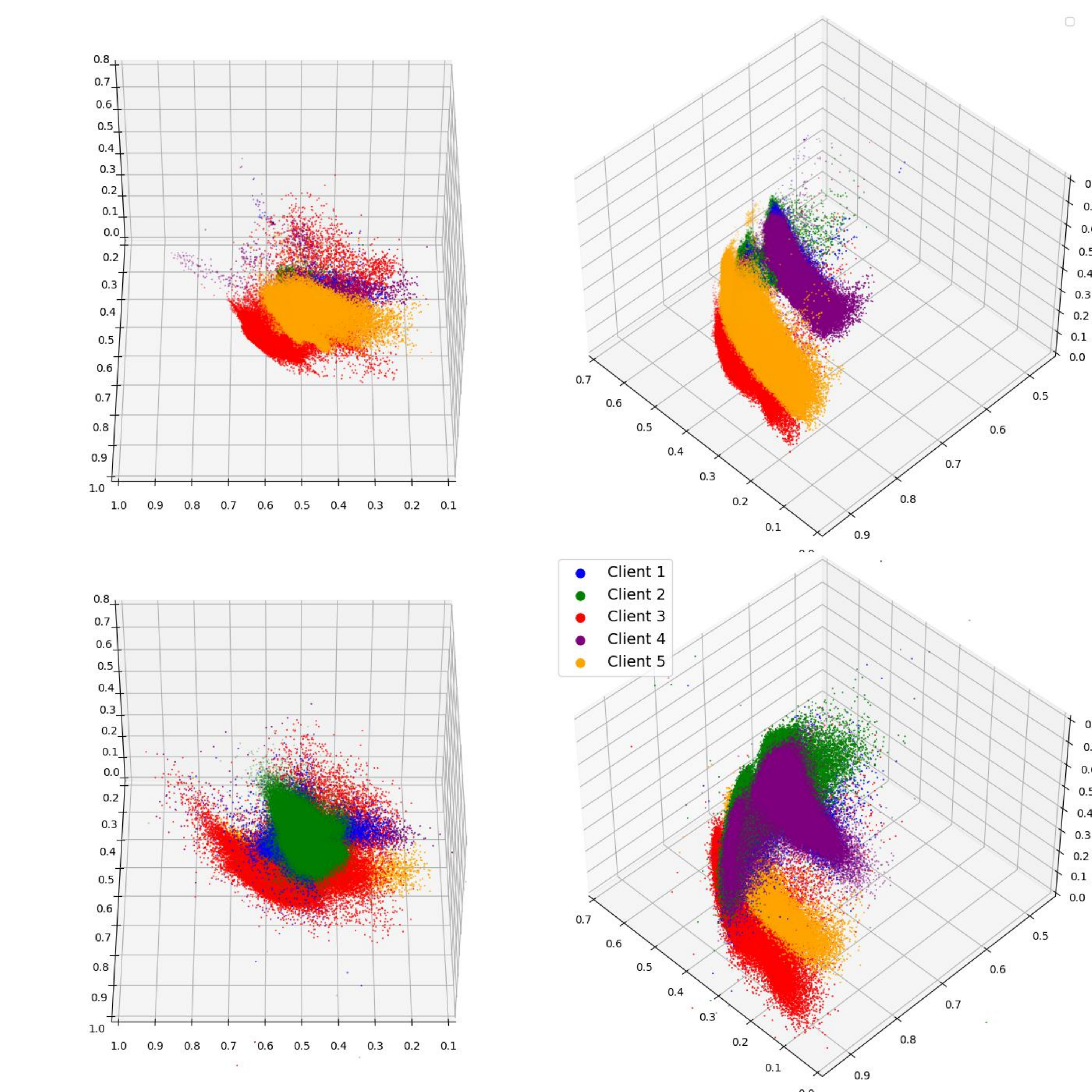}
        \caption{Stain matrix visualizations of the CAMELYON$17$ dataset~\cite{8447230c17}. Left column: H stain. Right column: E stain. Top row: Decomposition from histopathological images. Bottom row: Generated from a diffusion model.}\label{fig:c17_stain_matrices}
    \end{figure*}

\begin{figure*}[ht]
        \centering\includegraphics[width=1\linewidth]{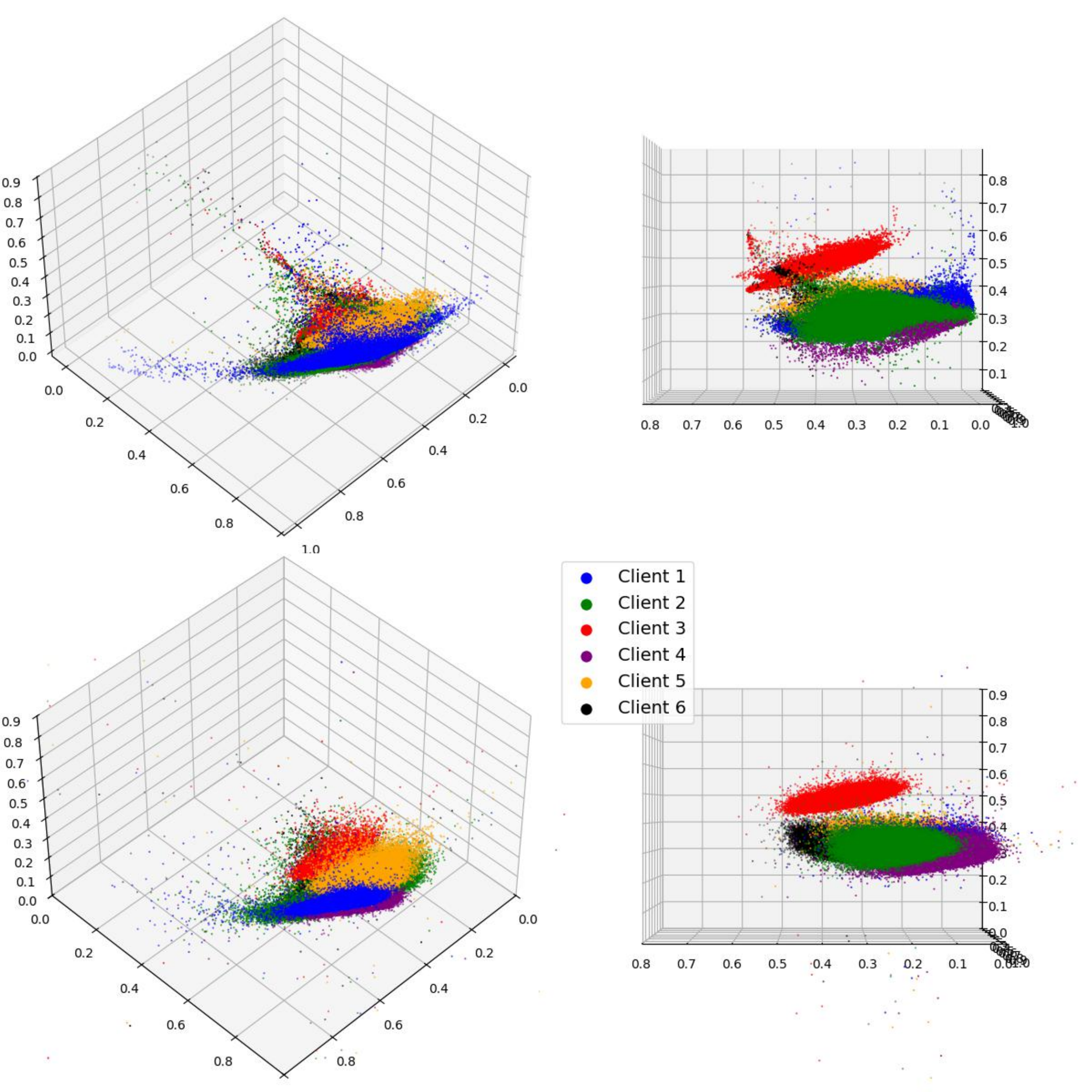}
        \caption{Stain matrix visualizations of the AGGC22 dataset~\cite{huo2024comprehensiveaggc}. Left column: H stain. Right column: E stain. Top row: Decomposition from histopathological images. Bottom row: Generated from a diffusion model.}\label{fig:a22_stain_matrices}
    \end{figure*}

    Second, we visualize stain matrices of each dataset, including those decomposed from histopathological patch images and those generated using the diffusion model, as shown in Figs.~\ref{fig:ma14_stain_matrices},~\ref{fig:c17_stain_matrices}, and~\ref{fig:a22_stain_matrices}.
    We observe that the stain matrices generated using the diffusion model for one client are almost identically distributed to those obtained from decomposing the histopathological patch images of the same client.

\begin{figure*}[ht]
        \centering\includegraphics[width=1\linewidth]{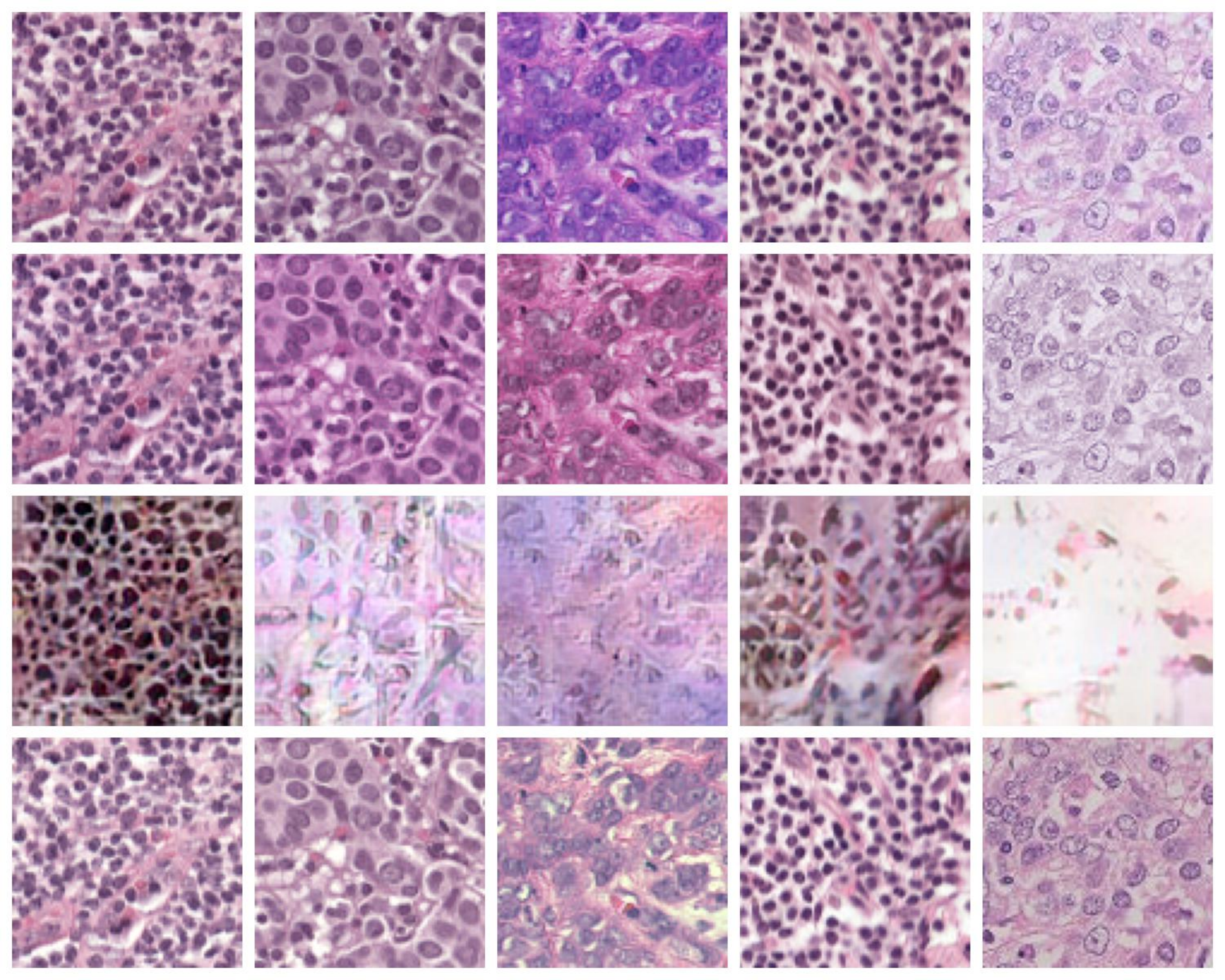}
        \caption{Sample images of the CAMELYON$17$~\cite{8447230c17} dataset, before and after applying FedSDA, CCST~\cite{chen2023federated}, and amp-norm. From left to right, it indicates from the 1st client to the 5th client.  Top row: Original images. Second row: Images with FedSDA applied. Third row: Images with CCST~\cite{chen2023federated} applied. Bottom row: Images with amp-norm applied.}\label{fig:c17_after_processing}
    \end{figure*}

\begin{figure*}[ht]
        \centering\includegraphics[width=1\linewidth]{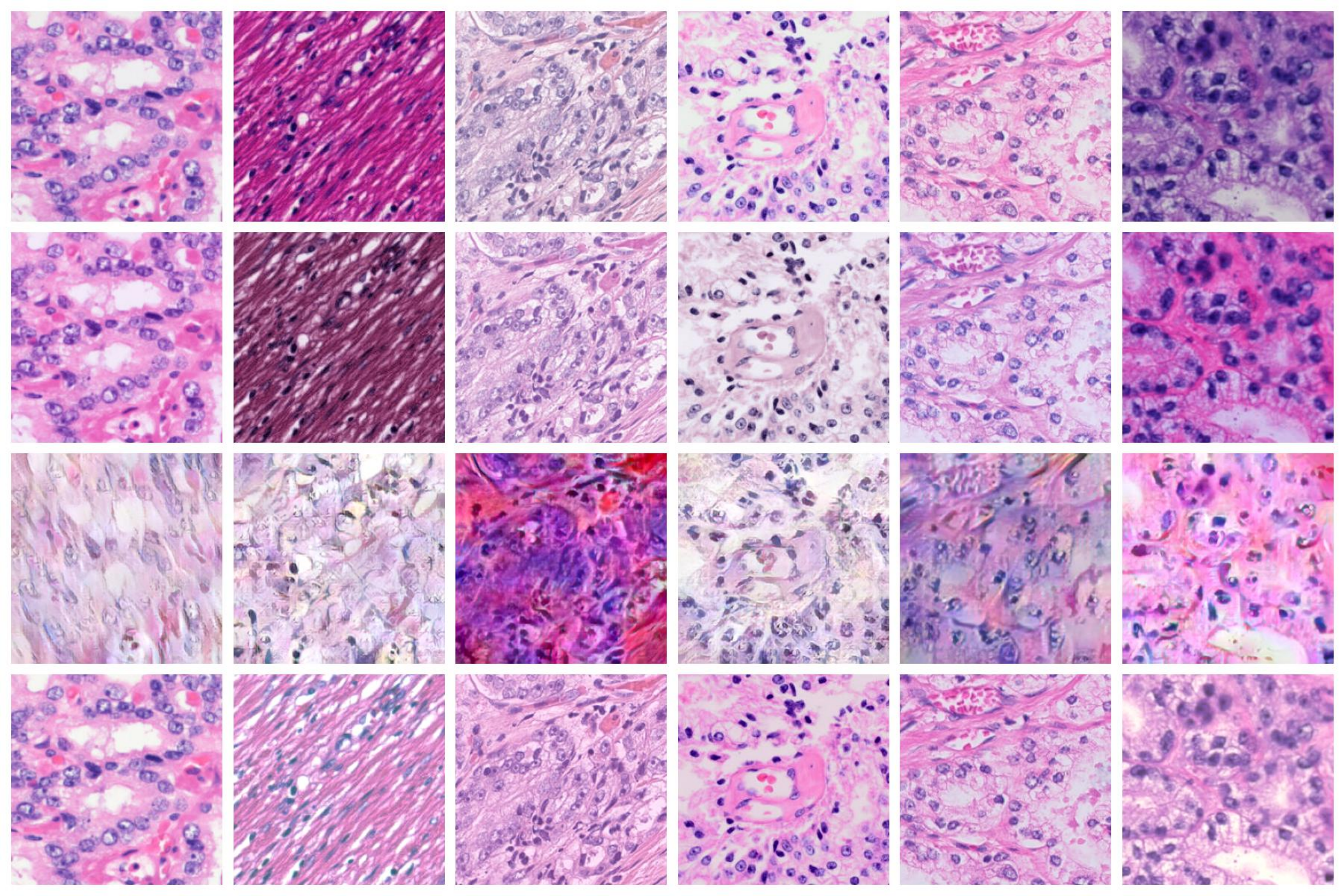}
        \caption{Sample images of the AGGC22 dataset~\cite{huo2024comprehensiveaggc},  before and after applying FedSDA, CCST~\cite{chen2023federated}, and amp-norm. From left to right, it indicates from the 1st client to the 5th client.  Top row: Original images. Second row: Images with FedSDA applied. Third row: Images with CCST~\cite{chen2023federated} applied. Bottom row: Images with amp-norm applied.}\label{fig:a22_after_processing}
    \end{figure*}

    Finally, we further present the sample images of each client in the C17 and A22 datasets after applying FedSDA, CCST~\cite{chen2023federated}, and amp-norm, respectively,  in Figs.~\ref{fig:c17_after_processing} and~\ref{fig:a22_after_processing}.
    From the results, we can observe that CCST~\cite{chen2023federated} significantly alters not only the stains of images but also their structures.
    In contrast, the structural alterations by amp-norm and FedSDA are not visually significant.
    On the other hand, across all clients, the stains in images with amp-norm applied are more monotone compared to those with FedSDA applied.


\end{document}